\documentclass[preprint,3p,times,12pt]{elsarticle}

\usepackage{amssymb}
\usepackage{subfigure}
\usepackage{float}
\usepackage{dsfont}
\usepackage{amsmath}
\usepackage[colorlinks]{hyperref}
\usepackage[x11names]{xcolor}
\usepackage{multirow}
\usepackage{booktabs}
\usepackage{graphicx}
\usepackage{breakurl}
\usepackage{algorithmic}
\usepackage[linesnumbered,ruled,vlined]{algorithm2e}

\AtBeginDocument{\hypersetup{linkcolor=red}}
\AtBeginDocument{\hypersetup{citecolor=green}}
\AtBeginDocument{\hypersetup{urlcolor=magenta}}

\journal{}

\begin{document}

\begin{frontmatter}

\title{Unified Learning Approach for Egocentric Hand Gesture Recognition and Fingertip Detection}

\author[label1]{Mohammad Mahmudul Alam}
\address[label1]{Department of Computer Science and Electrical Engineering, University of Maryland, Baltimore County, Baltimore, MD 21250, USA\fnref{label1}}
\ead{mdmahmudulalam007@umbc.edu}

\author[label2]{Mohammad Tariqul Islam}
\address[label2]{Department of Electrical Engineering, Princeton University, Princeton, NJ 08544, USA\fnref{label2}}
\ead{mtislam@princeton.edu}

\author[label3]{S. M. Mahbubur Rahman\corref{cor1}}
\address[label3]{Department of Electrical and Electronic Engineering, Bangladesh University of Engineering and Technology, Dhaka 1000, Bangladesh\fnref{label3}}
\cortext[cor1]{Corresponding author}
\ead{mahbubur@eee.buet.ac.bd}

\begin{abstract}
Head-mounted device-based human-computer interaction often requires egocentric recognition of hand gestures and fingertips detection. In this paper, a unified approach of egocentric hand gesture recognition and fingertip detection is introduced. The proposed algorithm uses a single convolutional neural network to predict the probabilities of finger class and positions of fingertips in one forward propagation. Instead of directly regressing the positions of fingertips from the fully connected layer, the ensemble of the position of fingertips is regressed from the fully convolutional network. Subsequently, the ensemble average is taken to regress the final position of fingertips. Since the whole pipeline uses a single network, it is significantly fast in computation. Experimental results show that the proposed method outperforms the existing fingertip detection approaches including the Direct Regression and the Heatmap-based framework. The effectiveness of the proposed method is also shown in-the-wild scenario as well as in a use-case of virtual reality.
\end{abstract}


\end{frontmatter}
\baselineskip 25pt

\section{Introduction}
In egocentric vision such as for virtual reality (VR), hand plays an instrumental role as a medium of interaction~\cite{b43}. The gesture of a hand and the location of its fingertips are essential information for a computer to understand the state of the interaction medium~\cite{singh2017trajectory}. Additionally, in VR environments, the recognition of hand gestures~\cite{b50}, and detection of fingertips~\cite{b47} are essential to interact between the virtual world and the real world. Existing hand gesture recognition and fingertip detection approaches can be broadly classified into two categories -~traditional image processing and \text{current} deep learning-based approaches. The early image processing approach relies mostly on the background segmentation and the shape and color of hand \text{in the so-called handcrafted algorithms}. Due to these dependencies, these methods often tend to fail in the presence of complex background, illumination effects, and in the variation of size and color of a person~\cite{gurav2015real}. 
\par 
On the contrary, the deep learning approach using convolutional neural networks (CNNs) has shown much better performance in these scenarios due to its capability to extract relevant features through learning algorithms. Since a given egocentric hand gesture has a given number of visible fingertips, traditional direct regression-based deep learning algorithms need to recognize hand gestures first, and afterward, they use corresponding trained fingertip detection model to detect the position of the fingertips~\cite{b14}. The problem arises since the number of visible fingers in a gesture can be variable but the number of outputs of a CNN must be fixed. Therefore, these algorithms require training different fingertip detection models for different hand gestures~\cite{b15}. 
\par
In this paper, we address this issue by proposing a unified approach to predict both the probabilistic output of the egocentric gesture of fingers and the positional output of all the fingertips using one forward propagation of a CNN. In the probabilistic output of gesture, the high probability indicates the existence of a visible finger while the low probability indicates the hidden finger. In general, the visible and hidden fingers are represented as labels `$1$' and `$0$', respectively. Hence, each gesture of hand can be recognized by the unique sequence of binary numbers by taking into account the probabilistic information of the fingers. Moreover, the proposed method estimates the coordinate position of fingertips by averaging the regressed ensemble of fingertip coordinates using a fully convolutional network (FCN), instead of using conventional direct regression using a fully connected (FC) layer. Thus, the estimation of the probability of fingers in a gesture and their relative sequence, and accurate positional information of fingertips make the overall hand gesture recognition and fingertip detection algorithm highly robust and reliable. Also, it is less likely to predict false positives and false negatives as compared to the existing direct regression~\cite{b37} and Heatmap-based~\cite{b26} frameworks. In particular, the proposed detection method results in significantly less pixel error as compared to the direct regression approach where pixel coordinates are directly regressed from an FC layer of a learning model. Besides, the proposed approach provides less localization error when compared to the Heatmap-based framework. In the following subsections, a literature review of previous works is presented and then the scope of analysis is given. Finally, specific contributions of this work are listed.

\subsection{Related Works}
Related works not only cover the egocentric hand gesture recognition, but also the generalized hand gesture recognition, because of the overlapping nature of the existing methodologies. For a detailed review of the field, the authors would like to suggest reading the survey paper by Bandini and Zariffa~\cite{b43}. Hand Gesture recognition and fingertip detection can be categorized into three different groups. The first group of works is concerned about gesture recognition. The second group of works is concerned with the detection of fingertips and the third group focuses on both gesture recognition and fingertip detection. The works on these groups are discussed in the following subsections.

\subsubsection{Gesture Recognition}
Hand gestures are mainly different combinations of fingers producing different shapes of a hand. Thus, the primary focus of gesture recognition methods that use image processing is shape matching or measuring dissimilarity among hand shapes. For instance, Ren~\emph{et al.}~\cite{b17} presented a part-based gesture recognition system that uses dissimilarity measure and template matching for an HCI application of arithmetic computation by gesture command. Discriminative 2D Zernike moments are also used for the recognition of static hand gestures of the ASL~\cite{b46}. In~\cite{b2}, CNN is used for hand gesture recognition in an HCI system, wherein the gesture is utilized to trigger mouse and keyboard events and to control a simulated robot. Lin~\emph{et al.}~\cite{b19} proposed that the background of a hand can be segmented first by using the Gaussian mixture model (GMM) and then the binarized image can be feed to a CNN classifier for learning instead of directly using the captured RGB image for hand gesture recognition. 
\par 
Different architectures of neural networks are applied for hand gesture recognition. Koller~\emph{et al.}~\cite{b13} embedded a CNN within an iterative expectation-maximization (EM) algorithm for the classification of hand shapes particularly in the case of continuous and weakly labeled data. Nunez~\emph{et al.}~\cite{b35} reported a method that combines the CNN and the long short-term memory (LSTM) network for skeleton-based temporal 3D hand gesture recognition. Xu~\emph{et al.}~\cite{xu2017hand} employed egocentric depth images for recognizing hand gesture or action.

\subsubsection{Fingertip Detection}
Image processing-based fingertip detection algorithms generally use background segmentation, contour analysis, and convex envelope techniques. Such a system is presented by Nguyen~\emph{et al.}~\cite{b11} where they first use a CNN-based hand detector, and then apply thresholding for hand segmentation in the detected region, and finally use the convex hull technique for fingertip detection. Deep learning-based fingertip detection mostly uses direct regression to predict the coordinate position of fingertips from the final FC layer of the CNN. However, Alamsyah~\emph{et al.}~\cite{b21} use an object detection algorithm by employing the region-based CNN (R-CNN) for predicting fingertips with an assumption that each fingertip is a class independent object. Huang~\emph{et al.}~\cite{b14} report a two-stage cascaded CNN-based direct regression for joint detection of fingertip and finger for a given hand gesture in egocentric vision. Similarly, Liu~\emph{et al.}~\cite{b15} use a bi-level cascaded CNN for detection of fingertips in a predetermined gesture in the egocentric videos. In the same vein, Huang~\emph{et al.}~\cite{b16} use two-stage CNN to detect fingertips from a hand image for an application of air writing wherein a fingertip acts like a pen. Jain~\emph{et al.}~\cite{b12} report the detection of only the index fingertip using a direct regression approach for a mixed-reality (MR) application in which the fingertip functions as a gestural interface for smartphones or head-mounted devices. Wetzler~\emph{et al.}~\cite{b44} mainly focus on CNN-based fingertip detection using a Kinect camera. This method uses a computationally extensive global orientation regression approach and an in-plane derotation scheme of depth images to predict the coordinate of fingertips.

\subsubsection{Gesture Recognition and Fingertip Detection}
An algorithm that detects a variable number of visible fingertips in a gesture implicitly recognizes the gesture too. For example, Prakash~\emph{et al.} \cite{b22} use a convex hull-based algorithm for detecting a variable number of visible fingertips, and hence, recognizing gesture concurrently for a limited HCI application. In contrast, Lai~\emph{et al.} \cite{b23} use two-step method for gesture recognition in a similar application setting. First, fingertips are detected using discrete curve evolution and then the gesture is recognized by partitioning the evolved curves detected from fingertips. Similarly, Meng~\emph{et al.}~\cite{b24} approximates the contours and convexity defect to find the coordinate positions of fingertips and then the gesture is recognized by using features such as the number of fingers, the Hu moments of a region bounded by the contour, and the compactness and the convexity of detected contour. Lee~\emph{et al.}~\cite{b9} estimates the scale-invariant angle between the fingers to determine the different number of visible fingertips. Afterward, fingertip gestures are recognized using a contour analysis of the fingers. Nguyen~\emph{et al.}~\cite{b27} use a deep learning-based approach where a modified multi-task segmentation network is employed for both segmentation of hand and detection of a variable number of fingertips. 
\par 
Wu~\emph{et al.} \cite{b26} represent the pixels of each fingertip as samples of 2D Gaussian distribution in the output tensor of Heatmap-based FCN in egocentric settings. By applying a suitable threshold, only the visible fingertips are detected that determines the gesture at the same time.

\subsection{Scope of Analysis}
Existing literature on egocentric hand gesture recognition and fingertip detection uses both image processing and deep learning-based approaches to confront the challenges. However, the image processing-based approaches have the dependency on background, hand shape and color thus tend to fail in complex and diverse scenarios. Moreover, the approaches that use the convex hull technique for gesture recognition and fingertip detection have their instinctive disadvantages. For instance, although they can recognize the gesture and detect fingertips, they cannot classify fingers and thus cannot apprise which fingertips have been detected. This prevents these methods from detecting correct gestures with positional information of fingertips. Consequently, we argue that deep learning-based detection will be more robust in diverse environments and finger classification. Nevertheless, deep learning-based direct regression approaches~\cite{b14} directly regress the fingertips in a predetermined gesture. So, there remains a scope of work in identifying hand gestures and finding fingertips concurrently. 
\par 
The direct regression approaches are simple, easy to implement, and require no post-processing. However, the CNN-based standard direct regression approach makes more pixel error as compared to the Heatmap-based methods. So, it is worthwhile to figure out a new way of direct regression approach that will result in less pixel error than the Heatmap-based solution with a slightly increased post-processing cost. Besides, Heatmap-~\cite{b26} and segmentation network-based~\cite{b27} approaches use a higher-order $(3^{\text{rd}})$ tensor representation which possesses complexity during post-processing. Hence, a unified gesture recognition and fingertip detection algorithm with a lower order $(1^{\text{st}}~\text{and}~2^{\text{nd}})$ tensor representation will reduce the post-processing complexity. Therefore, based on the motivations stated above, the development of CNN-based unified egocentric hand gesture recognition and fingertip detection algorithm is worth investigating.

\subsection{Specific Contributions}
In this paper, a CNN-based unified egocentric hand gesture recognition and fingertip detection algorithm is proposed for many potential applications in HCI. The specific contributions of the paper are as follows:

\begin{itemize}
\addtolength{\itemindent}{1mm}
\vskip 1.5mm 
\item A unified egocentric hand gesture recognition and fingertip detection algorithm using a lower order representation with a lower level of post-processing complexity is proposed

\vskip 1.5mm 
\item A new direct regression approach is introduced where an ensemble of fingertips position is directly regressed from FCN and later ensemble average is taken for the final position of fingertips

\vskip 1.5mm 
\item A higher level of accuracy in classification and a lower level of localization error in regression as compared to the well known direct regression and Heatmap-based framework is achieved through experimentations
\vskip 1.5mm 
\end{itemize}  

The rest of the paper is organized in the following order. In Section~\ref{sec:prop_method}, the proposed method is presented in detail. Section~\ref{sec:exps} includes the experiments and results along with a comparison with the existing methods and ablation study. Section~\ref{wild} shows the performance of the algorithm in the real-life images and a use-case of the method in the VR environment. Finally, Section~\ref{sec:conclusions} provides a conclusive remark.

\section{Proposed Method}\label{sec:prop_method}
The proposed method is a CNN-based unified egocentric hand gesture recognition and fingertip detection algorithm that combines the classification of gestures and regression of fingertips together. Using a single CNN both the probabilistic output for each of the fingers is predicted and the positional output of fingertips is regressed in one forward propagation of the network. In the following subsections, first, the unified detection algorithm is proposed, then CNN architecture for implementing the algorithm and fingertip detection method is explained. Finally, the optimization of the network is described.

\subsection{Unified Detection}
We unify the classification and regression into a single CNN using a lower-order binary representation. Hand gestures are the combination of different visible fingers where the total number of fingers in hand $N~(1~\leqslant~N~\leqslant~5)$ is fixed. However, in a specific gesture, the number of visible fingers $l~(l~\in~1,~2,~\cdots~,~N)$ is variable. Thus, for a specific gesture to locate the fingertips, the number of $x$-, and $y$-coordinates to be regressed from a CNN is $2l$. As the number of outputs of a CNN must be fixed and $l$ is variable here, we have addressed this issue by predicting the probabilistic output of length $N$ and regressing the positional output of length $2N$ from a single CNN. The probabilistic output is the binary representation of each finger, where `$1$' corresponds to the visible finger, and `$0$' corresponds to the finger being hidden. Consequently, each gesture will generate a unique sequence of binary numbers and from this sequence, the gesture can be recognized. Concurrently, as the binary sequence represents the visibility of fingers in a gesture, the positional output of the fingertips of the hidden finger can be set as \emph{don’t care} and ignored. Suppose, the probabilistic output of the CNN of length $N$ is $(p_1,~p_2,~\cdots~,~p_N)$ and the positional coordinate output of the CNN of length $2N$ is $((x_1,y_1),~(x_2,y_2),~\cdots~,(x_N,y_N))$ then the final output will be $(p_1~\times~(x_1,y_1),~p_2~\times~(x_2,y_2),~\cdots~,~p_N~\times~(x_N,y_N))$. From the final output, any $(0,0)$ coordinate will be considered as a hidden finger and ignored. If $(0,0)$ coordinate is considered as probable fingertip positional output, the probabilistic output can be further processed as $(2 p_n - 1)$ where $n~(n~\in~1,~2,~\cdots~,~N)$ to change the output range from $(0, 1)$ to $(-1, 1)$, and then only negative coordinates will be ignored.

\begin{figure}[!t]
\centering \subfigure[Example-1]
{\includegraphics[scale=.23]{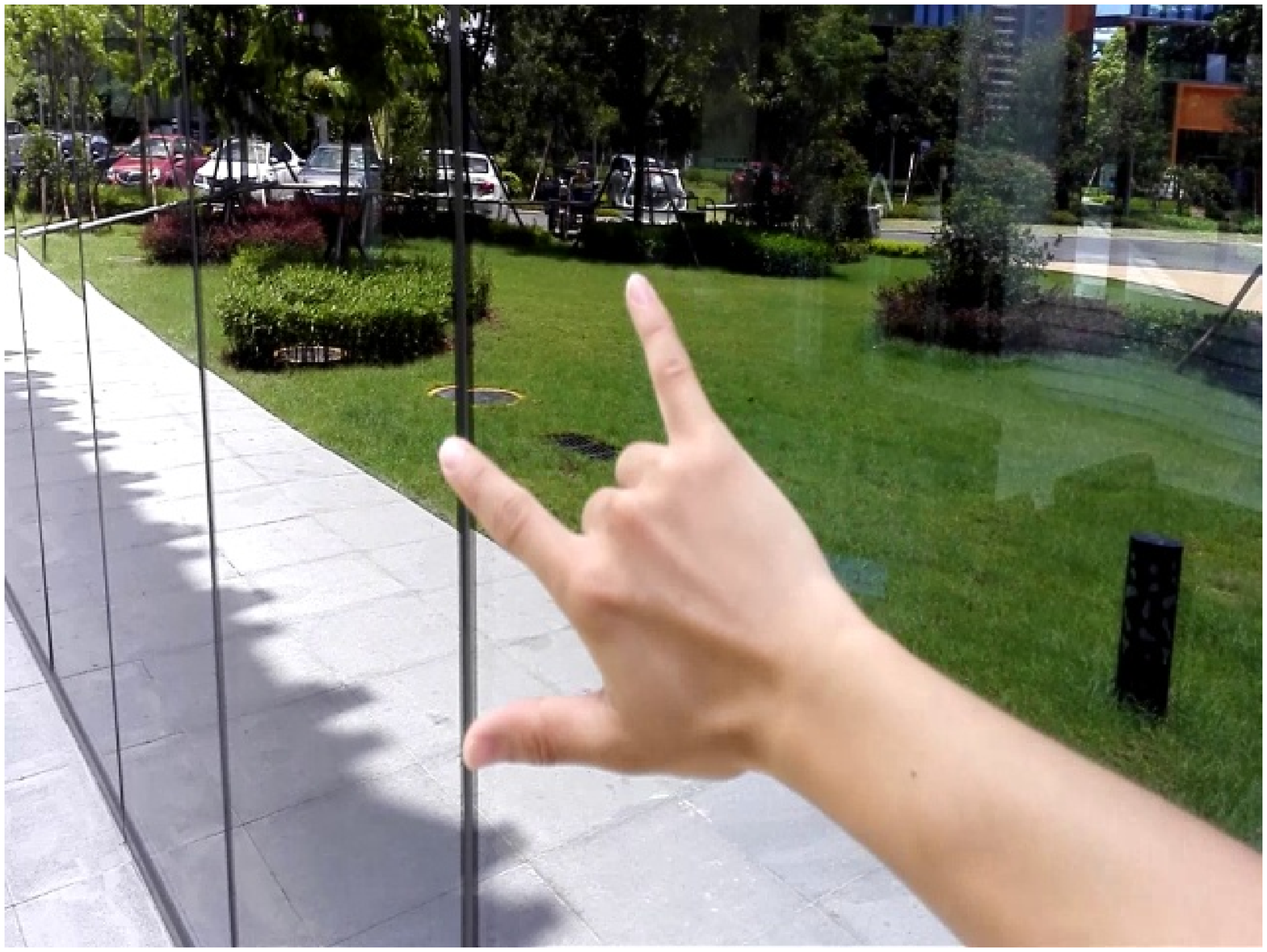}}
\hspace{5mm}
\subfigure[Example-2]
{\includegraphics[scale=.23]{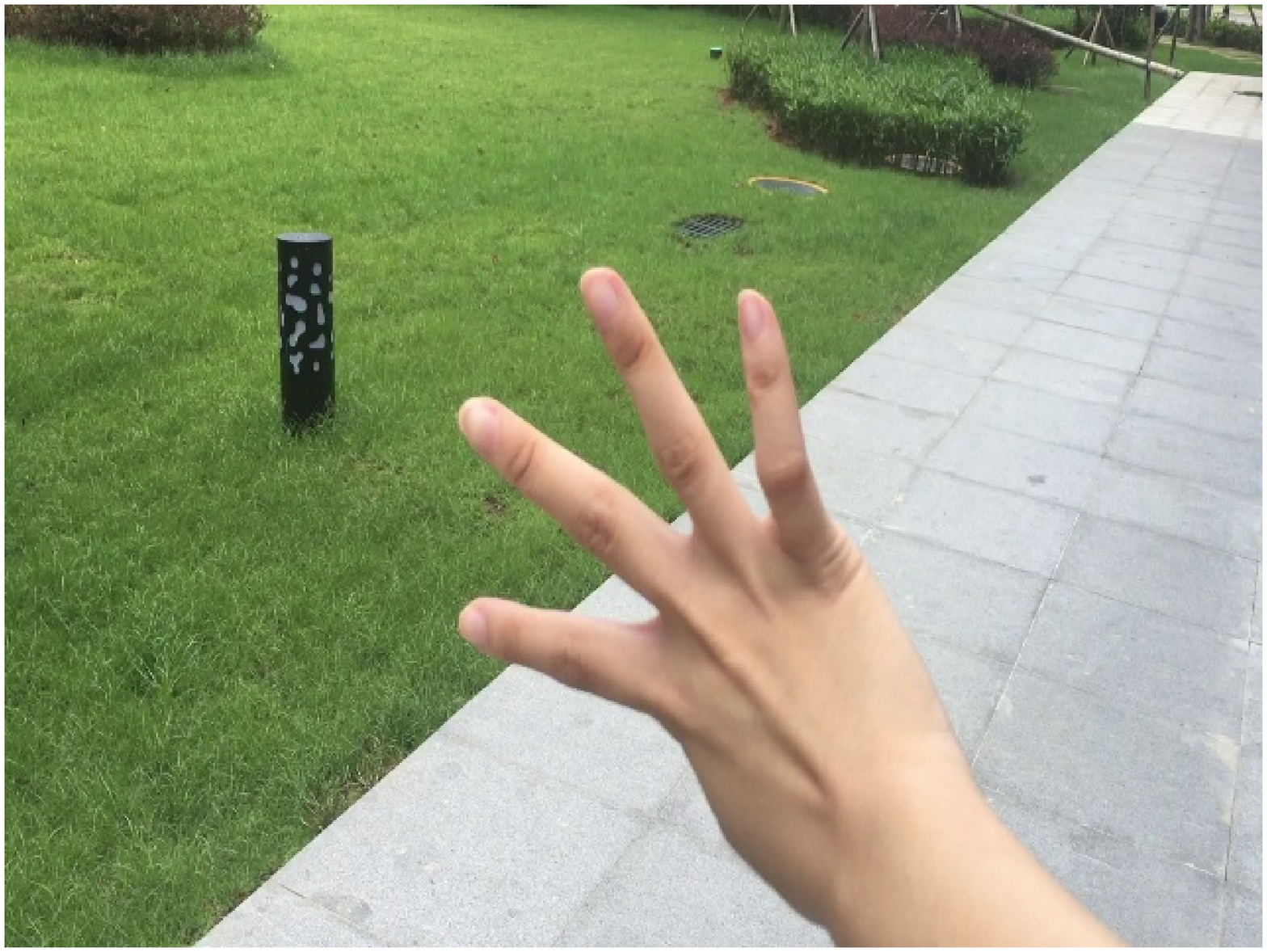}}  \caption{Illustrative images of the two different hand gestures are shown in (a) and (b).}
\label{exampleImages}
\end{figure}

Figure~\ref{exampleImages} shows two example images of hand gestures wherein Example-1 only thumb, index, and pinky fingers are visible and the middle and the ring fingers are hidden. So, the ground truth (GT) probabilistic binary output sequence for Example-1 will be $[1~1~0~0~1]$. Likewise, for Example-2 the GT probabilistic binary output sequence will be $[0~1~1~1~1]$. These are not only unique sequences for specific gestures but also apprise the visibility of the finger in a particular gesture which helps to determine which fingertip coordinates to ignore from the positional coordinates output of the CNN. 
\par 
During prediction, the probabilistic output will predict the visibility of fingers in a gesture. For a visible finger, it will give a higher confidence value and for a hidden finger, it will give a lower confidence value. So, a confidence threshold $\tau~(0<~\tau~<1)$ needs to be set above which the finger is visible and below which is hidden. Therefore, the criteria of detecting the visibility $p_n^{\prime}$ of fingers in a gesture from confidence value $p_n$ where $n~(n~\in~1,~2,~\cdots~,~N)$ is given by
\begin{equation}
p_n^{\prime}~=~\begin{cases}
1, & p_n > \tau \\
0, & p_n < \tau
\end{cases}
\end{equation}
For positional output, we propose an ensemble of direct regression from FCN where an ensemble of fingertips coordinates is regressed at first, and then the ensemble average is taken for final positional output of length $2N$ (both $x$-, and $y$-coordinates of $N$ fingers). Here, the ground truth ensemble of positional output is generated by stacking the same ground truth positional output $2N$ times for training purposes. The idea behind the stacking of the same output and creating an ensemble of positional output is that each output of the regression from the FCN will correspond to the different input features of the previous layer. Whereas, each output of the FC layer corresponds to all the input features of the previous layer. As a result, the output from the FCN will be more independent of a particular feature, and it is expected that even if few outputs may deviate from the ground-truth value which will be mitigated after taking the ensemble average. Therefore, a matrix $\mathbb{X}$ of size $2N\times2N$ is regressed at first from FCN, and then column-wise ensemble average is taken as the final output of fingertips position $\widetilde{\mathbb{X}}$ given by
\begin{equation}
\widetilde{\mathbb{X}}~=~\frac{1}{2N}~\sum_{i=1}^{2N}\mathbb{X}(:,~i)
\end{equation}

\subsection{CNN Architecture Design}
\begin{figure}[!t]
\centerline{\includegraphics[scale=.76]{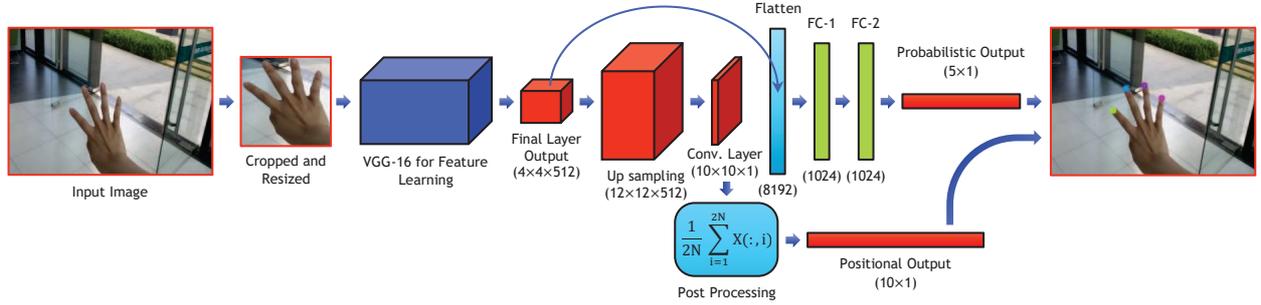}} \caption{A block diagram of the unified gesture recognition and fingertip detection algorithm depicting the CNN architecture with input and output.}\label{network}
\end{figure}

For gesture recognition and fingertip detection, the relevant portion of the hand is cropped from the input image using a bounding box and resized to $(128\times128)$. The resized image is used as the input to the proposed network for learning. During detection, the real-time object detection algorithm `you only look once’~(YOLO)~\cite{b36} is used for hand recognition in the first stage. Later, that hand portion can be cropped and resized to feed to the proposed framework. For feature learning, 16-layers visual geometry group (VGG) configuration given in~\cite{b28} is employed. This output is utilized to generate both the probabilistic output and positional output. First, the output of the feature learning stage is flattened and two FC layer is used back-to-back for better classification. Each of the FC layers is followed by a rectified linear unit (ReLU) activation function and a dropout layer. Finally, an FC layer is appended at the end to reduce the feature vector size to the same as that of the desired probabilistic output $\mathbb{P}$ of length $N$ given by
\begin{equation}
\mathbb{P} = \begin{bmatrix}
p_t & p_i & p_m & p_r & p_p
\end{bmatrix}^{\top}
\end{equation}
where from $p_t$ to $p_p$ are the probability of thumb ($t$), index ($i$), middle ($m$), ring ($r$), and pinky ($p$) finger, respectively. A sigmoid activation function is applied to the output of the final FC layer to normalize the probabilistic output. Moreover, the output of the feature learning stage is up-sampled followed by a ReLU activation function. Next, a convolution operation with a single filter is performed to further reduce the size of the feature vector to the same as that of the desired ensemble of positional output $\mathbb{X}$ of size $2N\times2N$ given by
\begin{equation}
\label{matrix}
\mathbb{X} = \begin{bmatrix}
x_t & y_t & x_i & y_i & x_m & y_m & x_r & y_r & x_p & y_p \\
x_t & y_t & x_i & y_i & x_m & y_m & x_r & y_r & x_p & y_p \\
x_t & \cdots & & & & & & & \cdots & y_p \\
\vdots & \ddots & & & & & & & \ddots & \vdots \\
x_t & \cdots & & & & & & & \cdots & y_p \\
\end{bmatrix}
\end{equation}
where $x_f$ and $y_f$~$(f~\in~t,~i,~m,~r,~p)$ stand for the coordinate position of the fingertips from thumb to pinky finger successively. In the final convolution operation, a linear activation function is applied. Finally, the column-wise ensemble average is taken as the final output of the fingertip positions. The overall system with CNN architecture is presented in Figure~\ref{network}. The activation functions and dropout layers are not shown in the figure for brevity.

\subsection{Optimization}
In the proposed framework, the probabilistic output and the positional output need to be optimized independently at the same time and thus two loss functions are defined. The probabilistic output predicts the binary sequence of `$1$' and `$0$' considering the visibility of the finger, and therefore, the following binary cross-entropy loss function is proposed to optimize the probabilistic output given by
\begin{equation}
\mathcal{L}_1 = \frac{1}{NM} \sum_{j=1}^{M} \, \sum_{k=1}^{N} {-\, \{\, \mathbb{P}_{(j\,k)}
\log_e{\hat{\mathbb{P}}_{(j\,k)}}} \, + \, {(1 - \mathbb{P}_{(j\,k)}) \times \log_e{(1 - \hat{\mathbb{P}}_{(j\,k)})}\,\}}
\end{equation}
where $N$ and $M$ represent the length of the probabilistic output and batch size, respectively. This loss function is the average of the loss over the batch.
\par 
The positional output regresses the ensemble of fingertips coordinate position which is a matrix of size $(2N\times2N)$. To optimize the positional output, the following mean squared error (MSE) loss function is proposed given by
\begin{equation}
\mathcal{L}_2 = \frac{1}{4N^{2}M} \sum_{j=1}^{{M}} \, \sum_{k=1}^{{2N}} \, \sum_{l=1}^{{2N}} \: \mathds{1}^{finger} \: \{\mathbb{X}_{(j\,k\,l)} - \hat{\mathbb{X}}_{(j\,k\,l)}\}^2
\end{equation}
where $\mathds{1}^{finger}$ denotes the visibility of the finger which is used for masking. If any finger is hidden in the gesture, the network should not be penalized for that fingertip regression. Hence, using the masking, fingertip detection loss for the hidden finger is eliminated. Finally, the total loss is the sum of the probabilistic and positional losses given by
\begin{equation}
\mathcal{L} = \mathcal{L}_1 + \mathcal{L}_2
\end{equation}
To optimize both of the loss functions $\mathcal{L}_1$ and $\mathcal{L}_2$, the commonly referred adaptive moment estimation (ADAM) optimizer is employed. This optimizer utilized the moving averages of both the first moment $m_k$ and second moment $v_k$ of the gradient of the loss functions that are given by~\cite{b29} 
\begin{equation}
m_k = \beta{_1} \times m_{k-1} + (1 - \beta{_1}) \times \Bigg\{\frac{d(\mathcal{L}_q)_k}{dw_k}\Bigg\}
\end{equation}
\begin{equation}
v_k = \beta{_2} \times v_{k-1} + (1 - \beta{_2}) \times {\Bigg\{\frac{d(\mathcal{L}_q)_k}{dw_k}\Bigg\}}^2
\end{equation}
where $q~(q~\in~1,~2)$,~$\beta_1$ and $\beta_2$~$(0<\beta_1,\beta_2<1)$ are the two hyper-parameters that control the decay rate of the moving averages, and $k$ stands for a particular iteration. Finally, the update of the weights of the model is given by
\begin{equation}
w_k = w_{k-1} - \frac{\eta \: m_k}{\sqrt{v}_k + \epsilon}
\end{equation}
where $\eta$ $(\eta>0)$ is the learning rate and $\epsilon~(\epsilon>0)$ is a infinitesimal number used for avoiding zero division error.

\subsection{Detection}

\begin{algorithm}[!t]
\label{algo}
\DontPrintSemicolon
\KwIn{$image$, $models$}
\KwOut{$probability~\mathbb{P}^{\prime}$, $position~\widetilde{\mathbb{X}}$}

\tcc{$(x_{tl},~y_{tl})$: top-left, $(x_{br},~y_{br})$: bottom-right coordinates of hand bounding box}

$(x_{tl},~y_{tl}),~(x_{br},~y_{br})~\leftarrow$~$yolo(image)$

$cropped\_image~\leftarrow~image[y_{tl}:y_{br},~x_{tl}:~x_{br}]$

$height,~width~\leftarrow~cropped\_image.shape$

$probability~\mathbb{P},~position~\mathbb{X}~\leftarrow~proposed\_method(cropped\_image)$

\tcc{post-processing}

$p_n^{\prime}~\forall~p_n^{\prime}~\in~\mathbb{P}^{\prime}~\leftarrow~\begin{cases}
1, & p_n > \tau \\
0, & p_n < \tau
\end{cases}
~\forall~p_n~\in~\mathbb{P}$

$\widetilde{\mathbb{X}}~\leftarrow~\frac{1}{2N}~\sum_{i=1}^{2N}\mathbb{X}(:,~i)$

$n~\leftarrow~0$~\tcc{index for probability}

\SetKw{KwBy}{by}
\For{$i=1$ \KwTo $2N$ \KwBy $2$}
{   
	\If {$\mathbb{P}^{\prime}[n]~==~1.0$}
	{
		\tcc{transforming coordinates to the original image}
		$\widetilde{\mathbb{X}}[i]~\leftarrow~\widetilde{\mathbb{X}}[i]~\times~width~+~x_{tl}$
		
		$\widetilde{\mathbb{X}}[i+1]~\leftarrow~\widetilde{\mathbb{X}}[i+1]~\times~height~+~y_{tl}$
	}
	
	$n~\leftarrow~n~+~1$ 
}

\caption{Unified Egocentric Hand Gesture Recognition and Fingertip Detection}
\end{algorithm}

During detection, in the first stage, the hand is detected using the YOLO object detection algorithm. Afterward, the detected hand portion from the image is cropped and resized to feed to the proposed network. The network predicts the probabilistic output of fingers and regresses the ensemble of fingertip positions. The probabilistic output of the network predicts a higher confidence value if the finger is visible and a lower confidence value if the finger is hidden in a gesture. To estimate a binary output sequence representing the array of visible fingers in hand, a confidence threshold $\tau$ is set. Due to the equal probability of the visibility or invisibility of the fingers, the confidence threshold $\tau$ is set to be $50\%$. As the proposed network directly regresses the ensemble of fingertip positional output $\mathbb{X}$, a column-wise ensemble average is estimated as the final fingertip positional output $\widetilde{\mathbb{X}}$. The entire step-by-step detection process is presented in Algorithm \ref{algo}.

\section{Experiments and Results}\label{sec:exps}
Experiments are performed based on the proposed method to validate the unified egocentric hand gesture recognition and fingertip detection algorithm. This section first presents the characteristics of the dataset on which experiments are carried out and a short description of data augmentation which is applied during the training period of the network. Afterward, the training and detection procedure of gesture recognition and fingertip detection are explained. Next, a short description of the comparing methods and performance metrics is provided. Finally, the results of the performance of the proposed approach are reported and compared with the existing methods which are presented both in terms of classification of hand gesture and regression of fingertips. All the training and testing code concerning the experimentations and results along with the pre-trained weights of the model are publicly available to download.~\footnote{\texttt{~Project:~\url{https://github.com/MahmudulAlam/Unified-Gesture-and-Fingertip-Detection}}}

\subsection{Dataset}
\begin{figure}[!t]
\centering
\subfigure[SingleOne]
{\includegraphics[scale=.168]{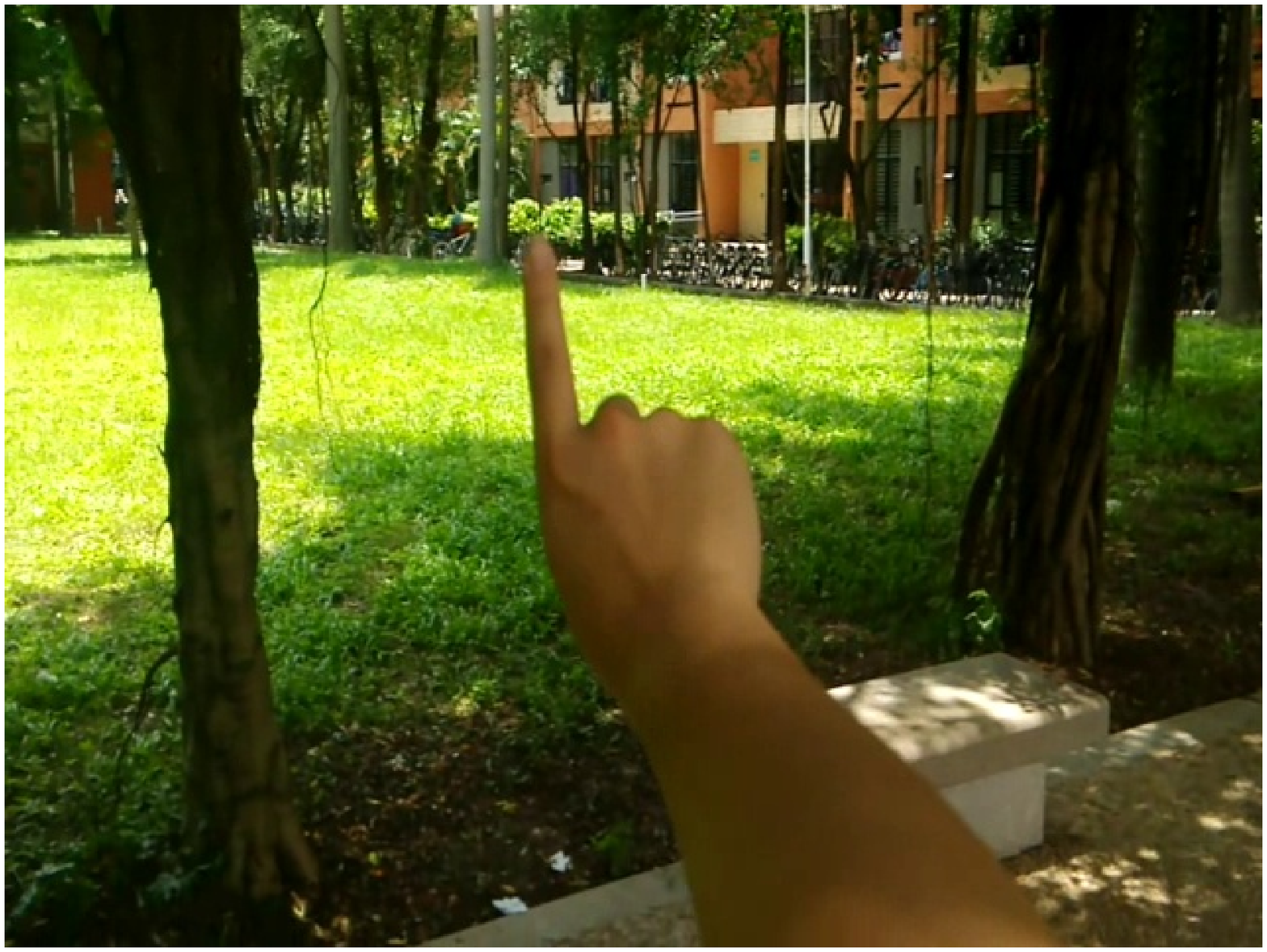}} \hspace{2mm}
\subfigure[SingleTwo] 
{\includegraphics[scale=.168]{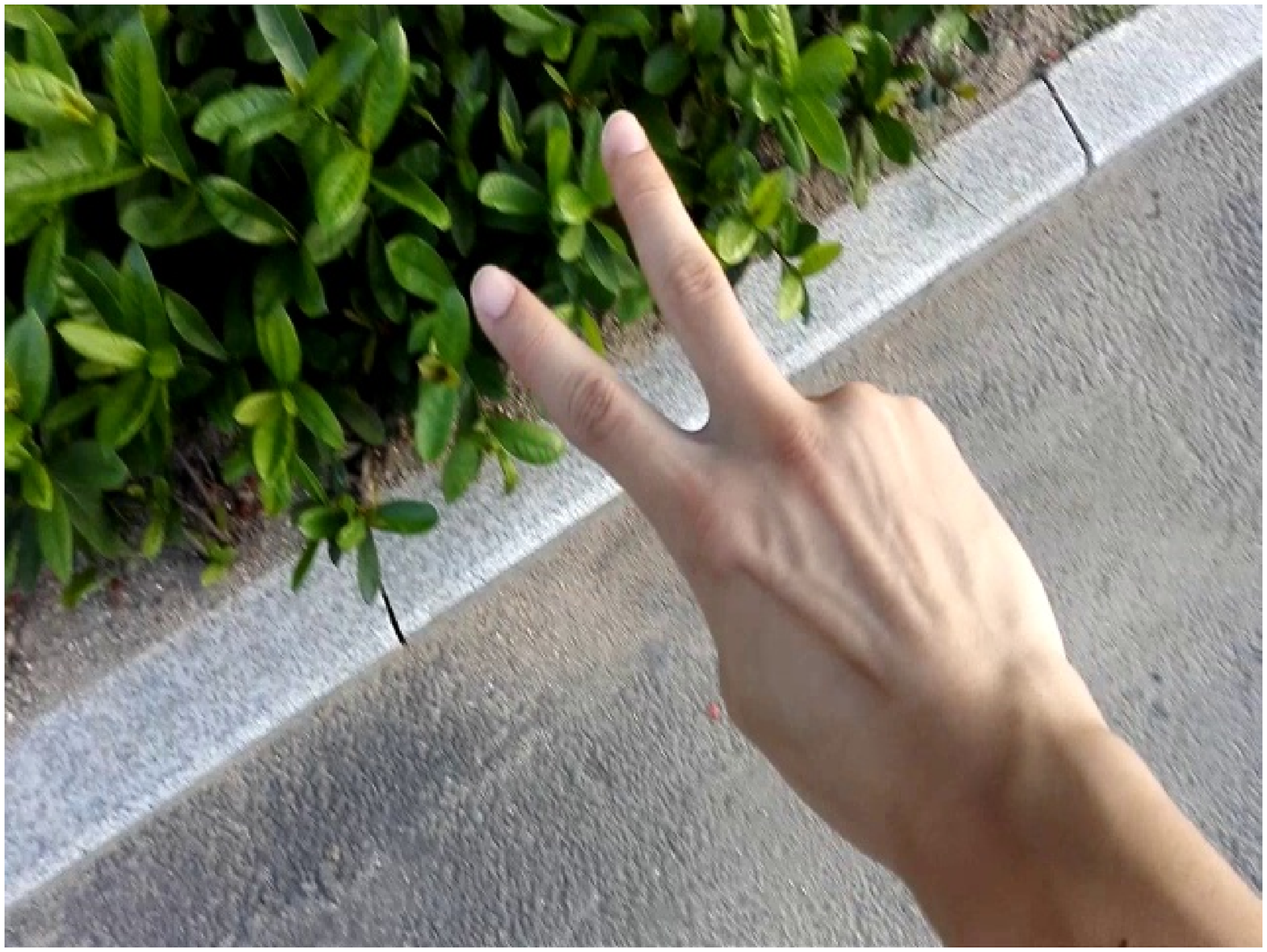}}
\hspace{2mm} \subfigure[SingleThree]
{\includegraphics[scale=.168]{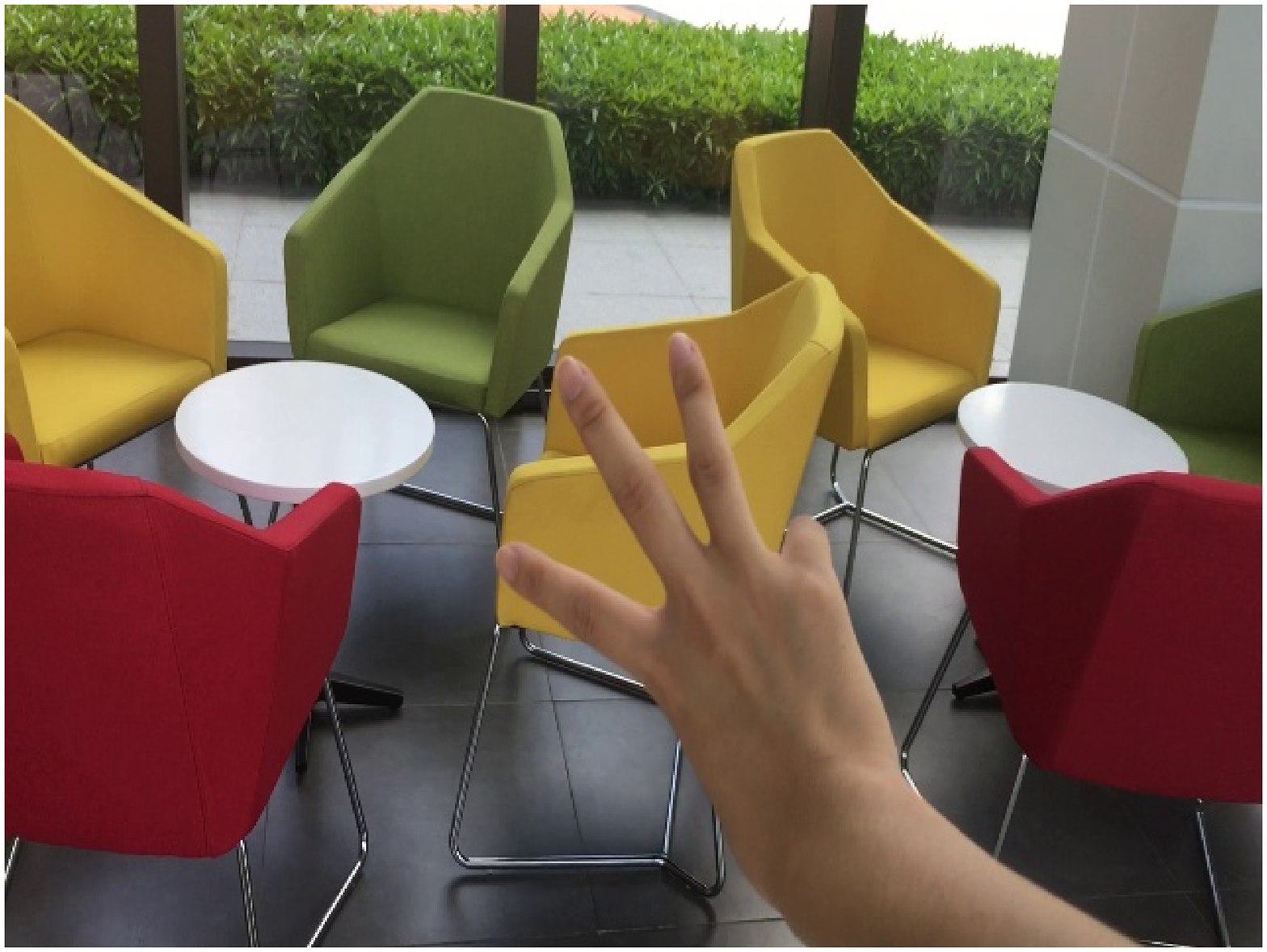}} \hspace{2mm}
\subfigure[SingleFour] {\includegraphics[scale=.168]{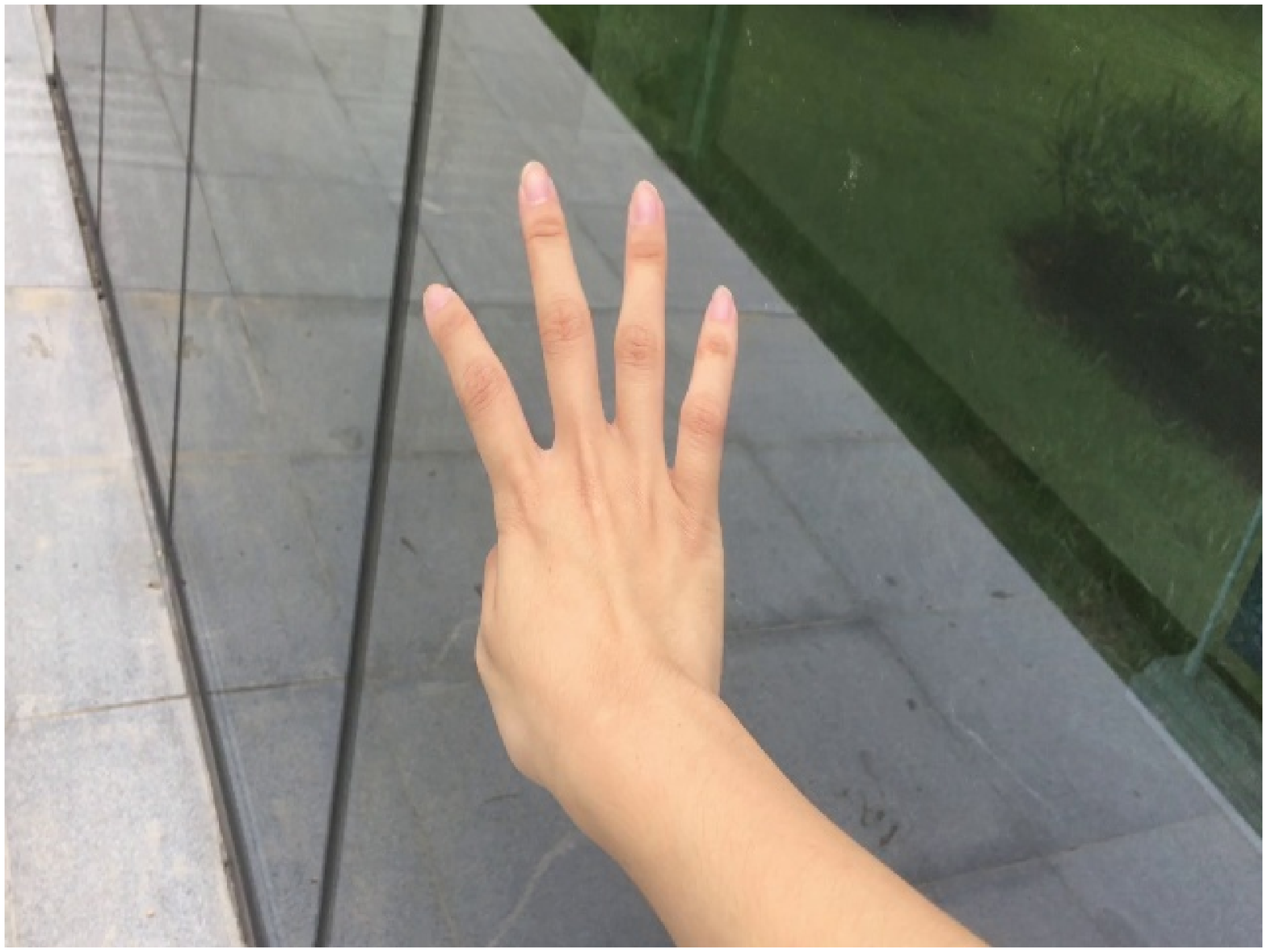}}
\hspace{2mm} \subfigure[SingleFive]
{\includegraphics[scale=.168]{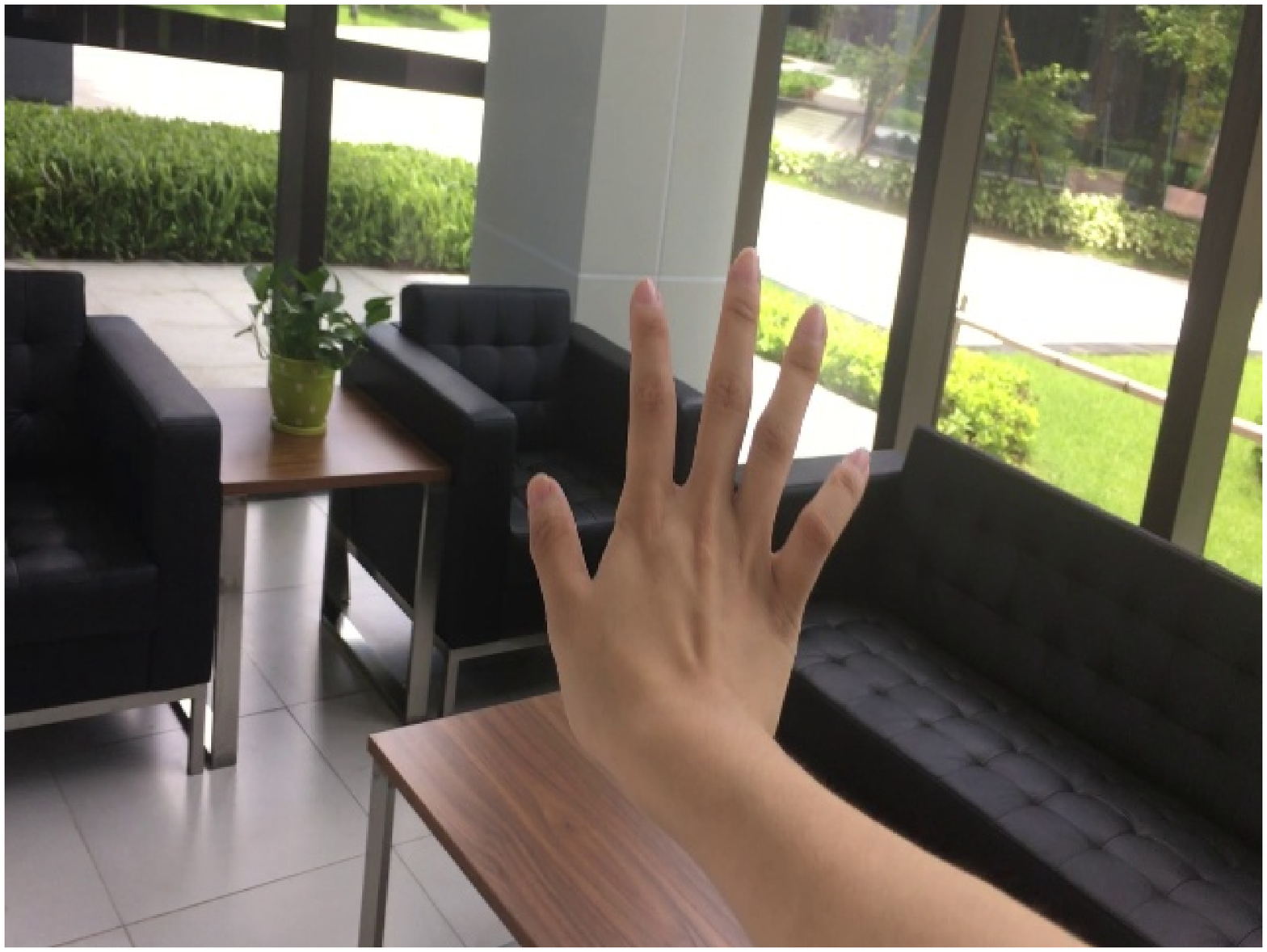}} \hspace{2mm}
\subfigure[SingleSix] {\includegraphics[scale=.168]{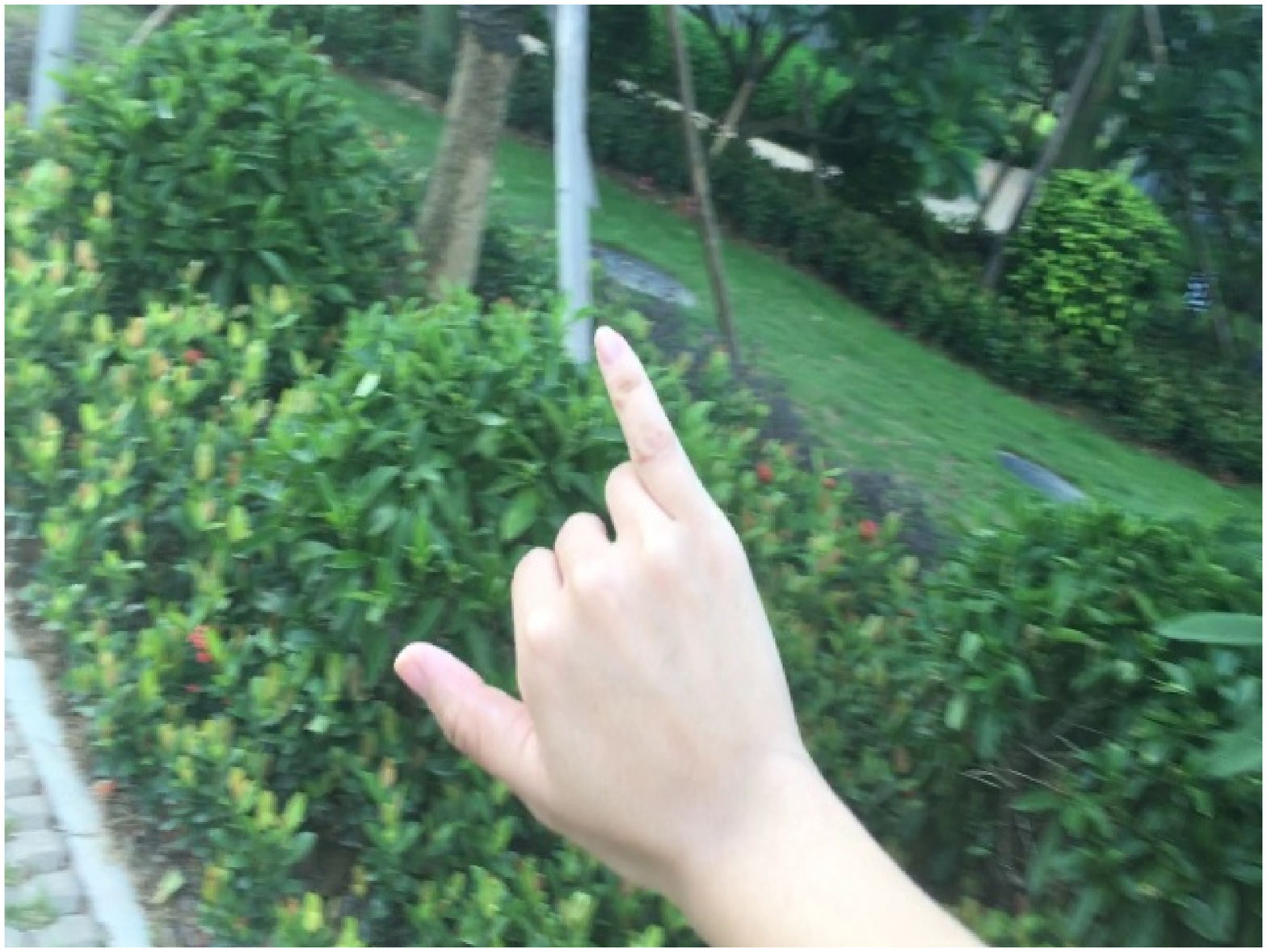}}
\hspace{2mm} \subfigure[SingleSeven]
{\includegraphics[scale=.168]{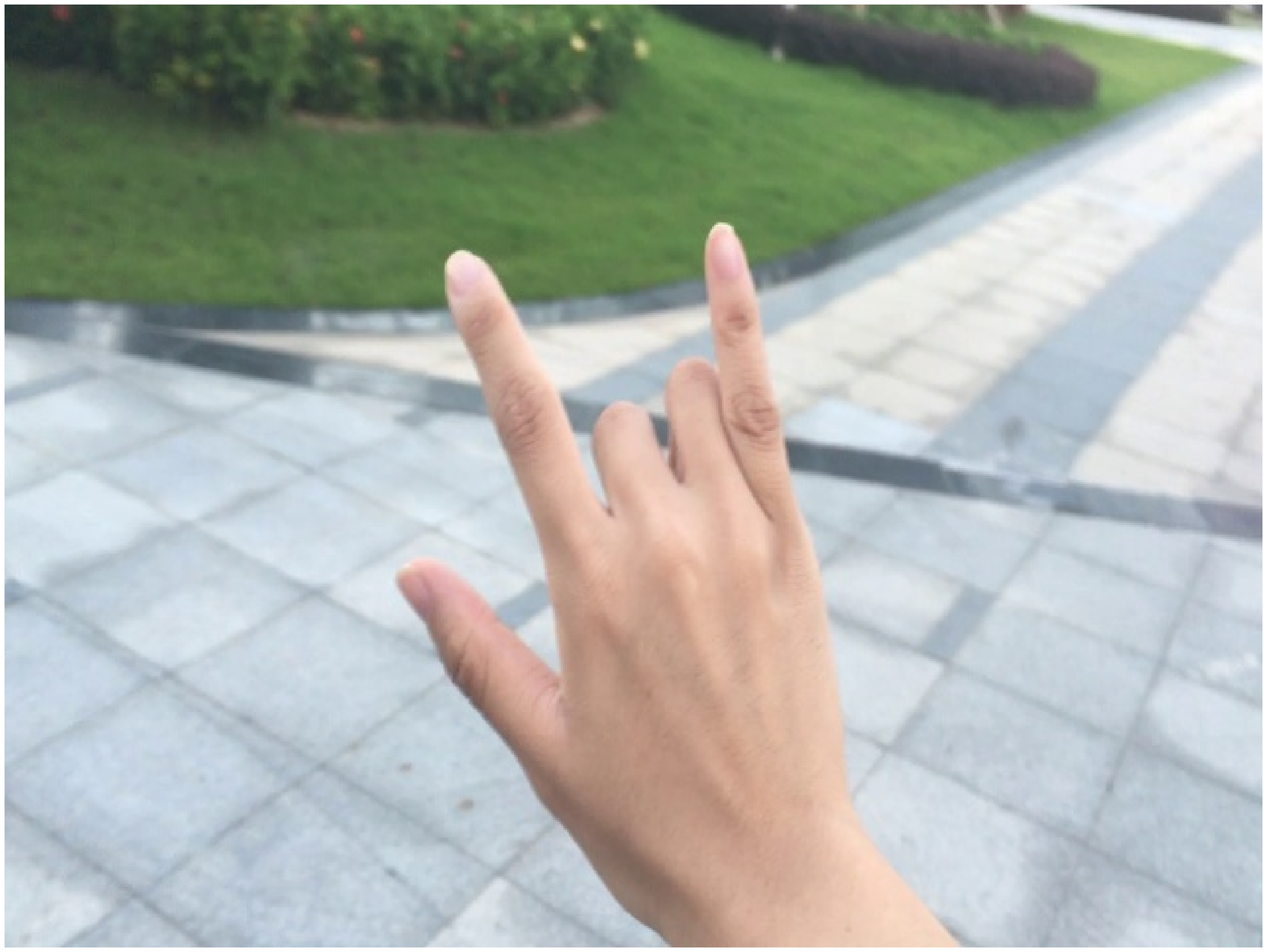}} \hspace{2mm}
\subfigure[SingleEight] {\includegraphics[scale=.168]{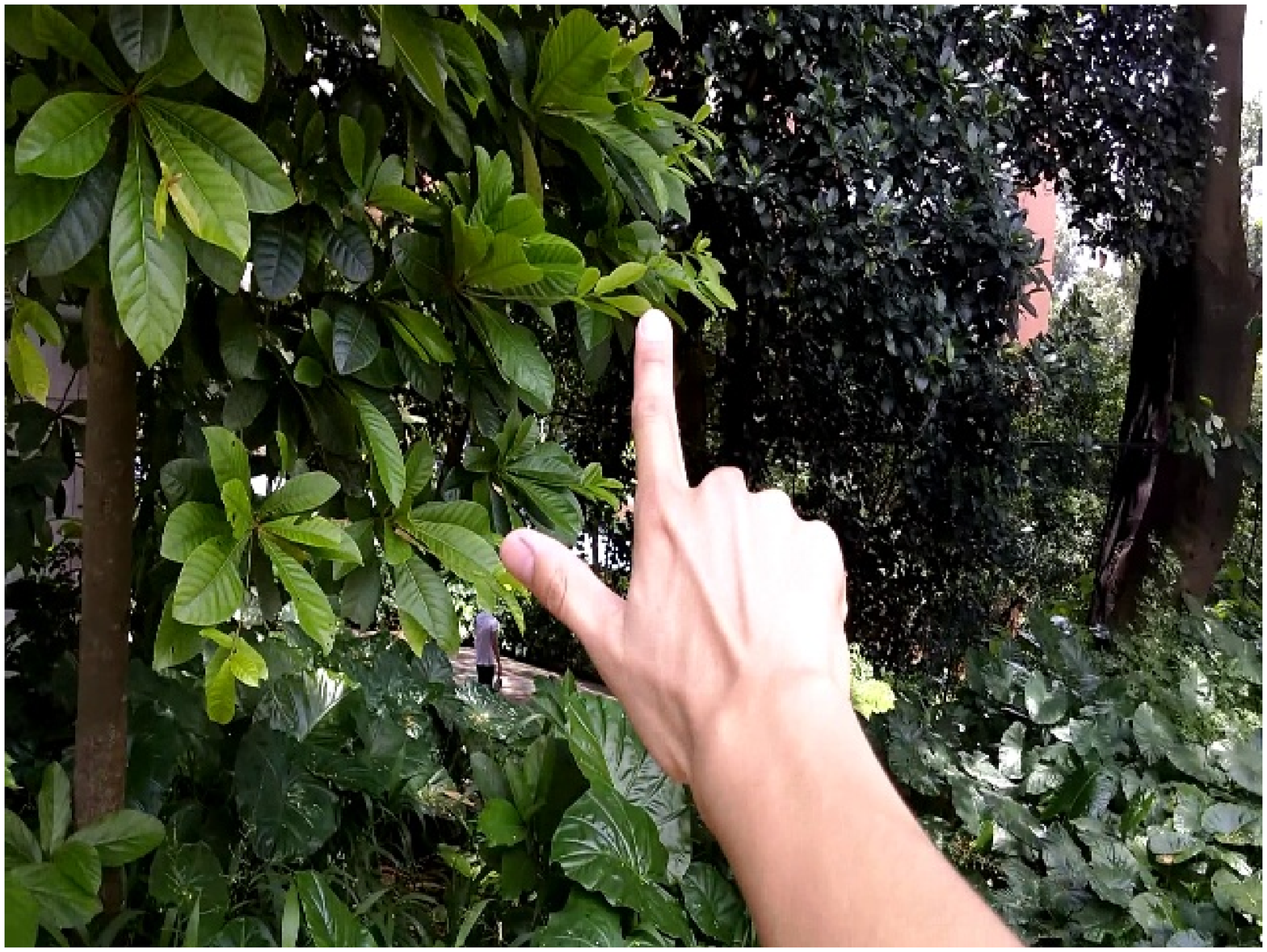}}
\caption{Visual examples of each of the eight gestures in the database are shown from (a) to (h).} 
\label{fig:dataset}
\end{figure}

In this experiment, the SCUT-Ego-Gesture database~\cite{b26} is employed for experimentation that contains eleven different datasets of single-hand gestures. Among these gesture datasets, eight are considered in the experimentation as they represent digit-type hand gestures. The eight datasets include $29,337$ RGB hand images in the egocentric vision each having a resolution of $640\times480$. Each of the datasets is partitioned into the test, validation, and training sets. First, for the test set $10\%$ images of each of the datasets are taken by randomly sampling one every ten images. Next, for the validation set $5\%$ images of the remaining images of the datasets are used by randomly sampling one every twenty images. Finally, the rest of the images of the datasets are employed for the training set. The number of images utilized in the test, validation, and training sets of different gesture classes are provided in Table~\ref{dataset}. Figure~\ref{fig:dataset} shows visual examples of hand gesture images of different classes where each gesture is constituted by a variable number of fingers. The list of names of the images used for the test, validation, and the training sets is made publicly available.~\footnote{\texttt{~Dataset:~\url{https://github.com/MahmudulAlam/Unified-Gesture-and-Fingertip-Detection/tree/master/dataset}}}

\begin{table}[!t]
\centering 
\caption{The list of the number of images used in the test, validation, and the training sets of the generic database}
\vspace{3mm}
\renewcommand{\arraystretch}{1.35}
\begin{tabular}{|c|c|c|c|c|}
\hline
Gesture Class & Test Set & Validation Set & Training Set &  Total\\ \hline
SingleOne & 337 & 151 & 2886 & 3374 \\ \hline
SingleTwo & 376 & 169 & 3218 & 3763 \\ \hline
SingleThree & 376 & 169 & 3223 & 3768 \\ \hline
SingleFour & 376 & 169 & 3222 & 3767 \\ \hline
SingleFive & 375 & 169 & 3211 & 3755 \\ \hline
SingleSix & 375 & 169 & 3213 & 3757 \\ \hline
SingleSeven & 377 & 169 & 3227 & 3773 \\ \hline
SingleEight & 338 & 152 & 2890 & 3380 \\ \hline
Total & 2930 & 1317 & 25090 & 29337 \\ \hline
\end{tabular}
\label{dataset}
\end{table}

\subsection{Data Augmentation}
To reduce the risk of overfitting, data augmentation is applied during training by including new training images artificially generated from the existing images of the datasets. In particular, the on-the-fly data augmentation process is used that generates new training images by applying random rotation, translation, shear transformation, illumination variation, scaling, cropping, additive Gaussian noise, and salt noise. The augmented images are generated randomly in each batch. As a result, the trained gesture recognition and fingertip detection model is learned from a large dataset. Hence, the trained model is expected to be generalized.

\subsection{Training}
To train the proposed gesture recognition and fingertip detection model, the relevant ground truth portion of the hand from the input image is cropped and resized to $(128\times128)$ using bilinear interpolation which is the input of the CNN. The model predicts a probabilistic output vector $\mathbb{P}$ of length $5$ and regresses an ensemble of positional output matrix $\mathbb{X}$ of size $(10\times10\times1)$. To generate the outputs of the desired size, the output tensor of the VGG-16 feature learning stage of size $(4\times4\times512)$ is flattened to a vector of length $8192$. The output vector length of the FC layers is chosen to be $1024$ and the dropout rate to be $0.5$. The final FC layer having an output length of $5$ is used to generate the probabilistic output. To produce the ensemble of the positional output of fingertips, the output tensor of the feature learning stage is three times up-sampled to $(12\times12\times512)$. Next, this output is convolved with a single filter of size $(3\times3)$ that results in a matrix of desired output size $(10\times10\times1)$. The proposed network is trained for a total of 300 epochs where the learning rate is lowered from $10^{-5}$ to $10^{-7}$ in a step by step process for better convergence. The parameter of the ADAM optimizer $\beta_1, \beta_2,$ and $\epsilon$ is chosen to be $0.9, 0.999,$ and $10^{-10}$, respectively, with a batch size of 64.

\begin{figure}[!htbp]
\centering 
\subfigure[Convergence of probabilistic loss function $\mathcal{L}_1$] 
{\includegraphics[scale=0.172]{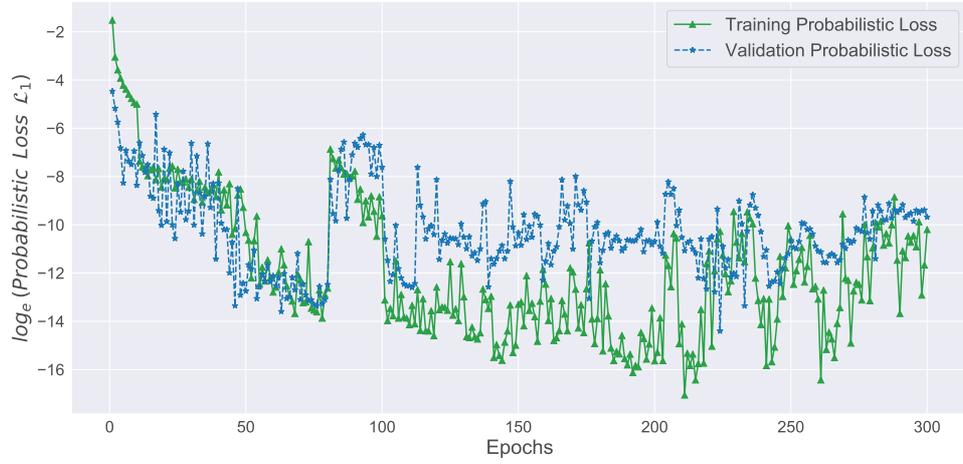}
\label{fig:prob_loss}} 
\subfigure[Convergence of positional loss function $\mathcal{L}_2$]
{\includegraphics[scale=0.172]{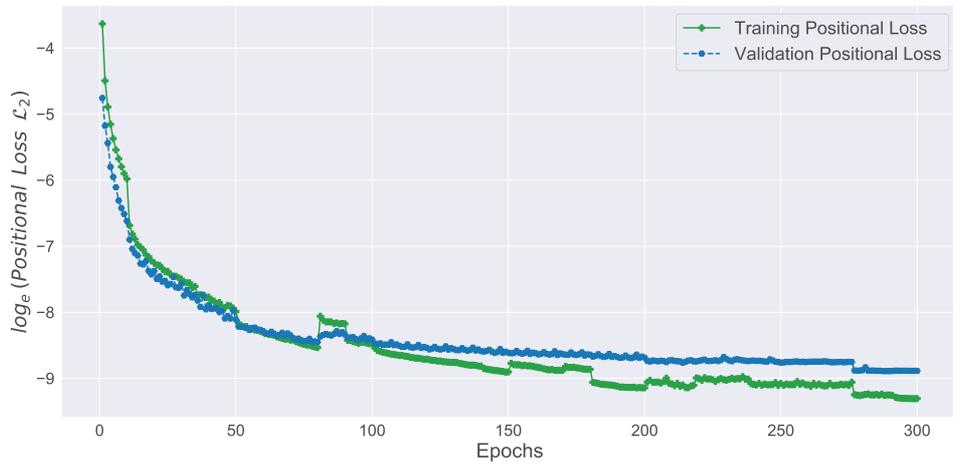} \label{fig:pos_loss}}
\subfigure[Learning curves in terms of the total loss $\mathcal{L}$] 
{\includegraphics[scale=0.172]{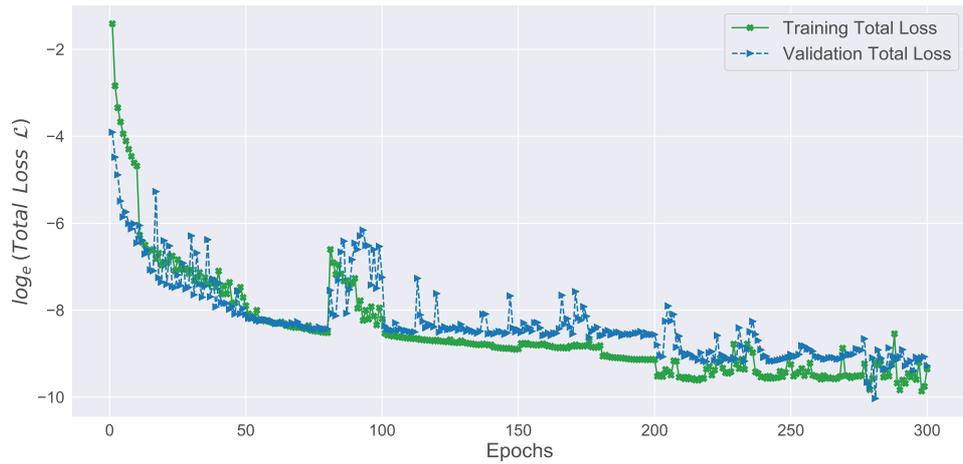}
\label{fig:total_loss}} 
\caption{The learning curves of the proposed unified gesture recognition and fingertip detection model. The convergence of the probabilistic, positional, and total loss functions are shown from (a) to (c), respectively.}
\label{fig:loss_curve}
\end{figure}

Figure~\ref{fig:loss_curve} shows the learning curves of the unified gesture recognition and fingertip detection model in terms of loss function both in the training and validation stages. Specifically, Figure~\ref{fig:prob_loss} shows the convergence of probabilistic loss function $\mathcal{L}_1$ and Figure~\ref{fig:pos_loss} shows the convergence of positional loss function $\mathcal{L}_2$. Figure~\ref{fig:total_loss} shows the learning curves in terms of the total loss $\mathcal{L}$ where the probabilistic and positional loss functions are combined. It can be seen from the learning curves that the proposed model is free from overfitting. During training, we have lowered the learning rate after few epochs and made changes in the augmentation, e.g., amount of rotation, translation, illumination, and so on. As a result, there are sudden fluctuations in the loss curves, especially visible in Figure~\ref{fig:prob_loss}. In a case, where a finger is hidden in the gesture, the positional loss is not penalized, and thus, we can observe less fluctuation as in Figure~\ref{fig:pos_loss}. As a matter of fact, modifying augmentation helped the model to be more robust and generalized and learn from a diverse dataset. In other words, the fluctuations in the learning curves reveal that the method avoids the problem of overfitting.

\subsection{Comparing Methods}
The proposed method is compared with the existing direct regression approach~\cite{b37} and the Heatmap-based gesture recognition and fingertip detection algorithm called `you only look what you should see' (YOLSE)~\cite{b26}. Before comparing to the proposed method, a brief description of these algorithms is provided here.

\begin{itemize}
\addtolength{\itemindent}{5mm} 
\vskip 1.5mm 
\item \emph{\textbf{YOLSE Approach:}} The YOLSE method of hand gesture recognition and fingertip detection algorithm is proposed by Wu~\emph{et al.}~\cite{b26} in 2017. They proposed a Heatmap-based approach using a fully convolutional network by representing each fingertip as a 2D Gaussian distribution in the output tensor. Each layer of the tensor represents a specific finger. The algorithm predicts a tensor and later from each layer of the tensor, the peak value is calculated. If the peak value exceeds a given threshold then the peak location is considered as the position of a visible fingertip. If the peak value falls below the threshold then that fingertip is considered hidden. 

\vskip 1.5mm 
\item \emph{\textbf{Direct Regression Approach:}} Mishra~\emph{et al.}~\cite{b37} proposed the direct regression-based hand gesture and fingertip detection algorithm in 2019. They employed MobileNetV2~\cite{b38} architecture as a backbone model and later produced a linear output using global average pooling. Afterward, from the same linear output, they used three fully connected (FC) layers for gesture classification, finger identification, and estimation of finger position. This algorithm is referred to as the Direct Regression approach as the final positional output of the fingertips are directly regressed from the FC layers.
\end{itemize}

\subsection{Performance Metrics}
The performance of the classification of egocentric hand gestures and that of estimation of the fingertips position are evaluated separately. The performance of the classification is assessed in terms of four measures, namely, accuracy, precision, recall, and F1 score. The higher the value of accuracy or F1 score, and the closer the value of precision or recall to unity, the better is the performance of the classification algorithm. In all of these evaluation metrics, unless otherwise stated, the confidence threshold is set to $50\%$. To evaluate the performance of estimation of fingertip position, the error in terms of mean Euclidean distance between ground truth pixel coordinate and regressed pixel coordinate is calculated as
\begin{equation}
\overline{D_f - \hat{D}_f}~=~\frac{1}{S \langle\mathbb{P}, \mathbf{1}\rangle}~\sum_{k=1}^{S} \sum_{j=1}^{\langle\mathbb{P}, \mathbf{1}\rangle} {(p_f^{\prime})}_{j\,k}~\sqrt{\{(x_f)_{j\,k} - (\hat{x}_f)_{j\,k}\}^2 +~\{(y_f)_{j\,k} - (\hat{y}_f)_{j\,k}\}^2}
\end{equation}
where $f~(f~\in~t,~i,~m,~r,~p)$, S stands for the total number of correctly recognized gestures in the test set in a particular class, and $\langle\mathbb{P}, \mathbf{1}\rangle$ is the number of total fingers in the gesture.

\begin{table}[!t]
\centering
\caption{Performance of gesture classification of the comparing methods in terms of Accuracy, Precision, Recall, and F1 score}
\vspace{3mm}
\label{tab:classification}
\resizebox{\textwidth}{!}{%
\renewcommand{\arraystretch}{2.0}
\begin{tabular}{|c|c|c|c|c|c|c|c|c|c|c|}
\hline
\multirow{2}{*}{Method} & \multirow{2}{*}{Metric} & \multicolumn{8}{c|}{Gesture} & \multirow{2}{*}{Mean} \\ \cline{3-10}
 &  & SingleOne & SingleTwo & SingleThree & SingleFour & SingleFive & SingleSix & SingleSeven & SingleEight &  \\ \hline

\multirow{4}{*}{\shortstack{GT-\\YOLSE}} & \textbf{Accuracy (\%)} & 96.72 & 98.16 & 98.46 & 97.99 & 98.87 & 99.08 & 98.74 & 96.86 & 98.11 \\ \cline{2-11}
 & \textbf{Precision (\%)} & 85.13 & 97.63 & 96.88 & 97.89 & 100.00 & 98.33 & 97.22 & 93.31 & 95.80  \\ \cline{2-11}
 & \textbf{Recall (\%)} & 86.65 & 87.77 & 90.96 & 86.17 & 91.20 & 94.40 & 92.84 & 78.40 & 88.55  \\ \cline{2-11}
 & \textbf{F1 Score} & 0.8588 & 0.9244 & 0.9383 & 0.9165 & 0.9540 & 0.9633 & 0.9498 & 0.8521 & 0.9196 \\ \hline \hline

\multirow{4}{*}{\shortstack{YOLO-\\YOLSE}} & \textbf{Accuracy (\%)} & 97.00 & 98.16 & 98.46 & 98.19 & 99.01 & 99.04 & 98.70 & 97.00 & 98.20 \\ \cline{2-11}
 & \textbf{Precision (\%)} & 86.94 & 96.26 & 96.62 & 98.50 & 100.00 & 98.87 & 97.75 & 93.71 & 96.08 \\ \cline{2-11}
 & \textbf{Recall (\%)} & 86.94 & 89.10 & 91.22 & 87.23 & 92.27 & 93.60 & 92.04 & 79.29 & 88.96 \\ \cline{2-11}
 & \textbf{F1 Score} & 0.8694 & 0.9254 & 0.9384 & 0.9252 & 0.9598 & 0.9616 & 0.9481 & 0.8590 & 0.9234 \\ \hline \hline

\multirow{4}{*}{\shortstack{GT-\\Direct\\Regression}} & \textbf{Accuracy (\%)} & 99.97 & 99.90 & 99.86 & 99.86
& 99.76 & 99.62 & 99.52 & 99.73 & 99.78 \\ \cline{2-11}
 & \textbf{Precision (\%)} & 100.00 & 99.73 & 99.47 & 99.47 & 99.46 & 98.66 & 98.40 & 97.60 & 99.10 \\ \cline{2-11}
 & \textbf{Recall (\%)} & 99.70 & 99.47 & 99.47 & 99.47 & 98.67 & 98.40 & 97.88 & 100.00 & 99.13 \\ \cline{2-11}
 & \textbf{F1 Score} & 0.9985 & 0.9960 & 0.9947 & 0.9947 & 0.9906 & 0.9853 & 0.9814 & 0.9879 & 0.9911 \\ \hline \hline

\multirow{4}{*}{\shortstack{YOLO-\\Direct\\Regression}} & \textbf{Accuracy (\%)} & 99.69 & 99.93 & 99.93 & 99.90
& 99.86 & 99.93 & 99.90 & 99.59 & 99.84 \\ \cline{2-11}
 & \textbf{Precision (\%)} & 97.95 & 99.47 & 99.73 & 99.47 & 100.00 & 99.73 & 99.73 & 98.80 & 99.36 \\ \cline{2-11}
 & \textbf{Recall (\%)} & 99.41 & 100.00 & 99.73 & 99.73 & 98.93 & 99.73 & 99.47 & 97.63 & 99.33 \\ \cline{2-11}
 & \textbf{F1 Score} & 0.9867 & 0.9973 & 0.9973 & 0.9960 & 0.9946 & 0.9973 & 0.9960 & 0.9821 & 0.9934 \\ \hline
\hline

\multirow{4}{*}{\shortstack{GT-\\Proposed\\Method}} &  \textbf{Accuracy (\%)} & 99.97 & 100.00 & 100.00 & 99.97
& 99.93 & 99.93 & 99.93 & 99.93 & \textbf{99.96} \\ \cline{2-11}
 & \textbf{Precision (\%)} & 99.70 & 100.00 & 100.00 & 99.73 & 100.00 & 100.00 & 99.47 & 99.70 & \textbf{99.82} \\ \cline{2-11}
 & \textbf{Recall (\%)} & 100.00 & 100.00 & 100.00 & 100.00 & 99.47 & 99.47 & 100.00 & 99.70 & \textbf{99.83} \\ \cline{2-11}
 & \textbf{F1 Score} & 0.9985 & 1.0000 & 1.0000 & 0.9987 & 0.9973 & 0.9973 & 0.9974 & 0.9970 & \textbf{0.9983} \\ \hline \hline

\multirow{4}{*}{\shortstack{YOLO-\\Proposed\\Method}} & \textbf{Accuracy (\%)} & 99.90 & 100.00 & 100.00
& 99.93 & 99.90 & 99.93 & 99.90 & 99.90 & 99.93 \\ \cline{2-11}
& \textbf{Precision (\%)} & 99.12 & 100.00 & 100.00 & 99.47 & 100.00 & 100.00 & 99.21 & 100.00 & 99.72 \\ \cline{2-11}
& \textbf{Recall (\%)} & 100.00 & 100.00 & 100.00 & 100.00 & 99.20 & 99.47 & 100.00 & 99.11 & 99.72 \\ \cline{2-11}
& \textbf{F1 Score} & 0.9956 & 1.0000 & 1.0000 & 0.9973 & 0.9960 & 0.9973 & 0.9960 & 0.9955 & 0.9972 \\ \hline 
\end{tabular}%
}
\end{table}

\begin{table}[!t]
\centering
\caption{Performance of fingertip positional accuracy of the comparing methods in terms of the mean pixel (px) error}
\vspace{3mm}
\label{tab:regression}
\resizebox{\textwidth}{!}{%
\renewcommand{\arraystretch}{1.8}
\begin{tabular}{|c|c|c|c|c|c|c|c|c|c|}
\hline
\multirow{2}{*}{Method} & \multicolumn{8}{c|}{Gesture} & \multirow{2}{*}{\shortstack{Mean\\Error\\(px)}} \\ \cline{2-9}
 & SingleOne & SingleTwo & SingleThree & SingleFour & SingleFive & SingleSix & SingleSeven & SingleEight &  \\ \hline

\multirow{2}{*}{\shortstack{GT-\\YOLSE}} & \multirow{2}{*}{5.71 $\pm$ 15.29} & \multirow{2}{*}{4.16 $\pm$ 3.89} & \multirow{2}{*}{3.51 $\pm$ 1.92} & \multirow{2}{*}{3.95 $\pm$ 4.76} & \multirow{2}{*}{3.74 $\pm$ 1.61} & \multirow{2}{*}{3.59 $\pm$ 1.56} & \multirow{2}{*}{3.89 $\pm$ 1.66} & \multirow{2}{*}{5.22 $\pm$ 2.48} & \multirow{2}{*}{4.22 $\pm$ 4.15} \\
 &  &  &  &  &  &  &  &  &  \\ \hline

\multirow{2}{*}{\shortstack{YOLO-\\YOLSE}} & \multirow{2}{*}{5.06 $\pm$ 9.53} & \multirow{2}{*}{4.31 $\pm$ 4.56} & \multirow{2}{*}{3.56 $\pm$ 2.20} & \multirow{2}{*}{3.6 $\pm$ 2.46} & \multirow{2}{*}{3.76 $\pm$ 1.65} & \multirow{2}{*}{3.62 $\pm$ 1.51} & \multirow{2}{*}{3.98 $\pm$ 2.68} & \multirow{2}{*}{5.14 $\pm$ 2.66} & \multirow{2}{*}{4.13 $\pm$ 3.41} \\
 &  &  &  &  &  &  &  &  &  \\ \hline

\multirow{2}{*}{\shortstack{GT-\\Direct\\Regression}} & \multirow{2}{*}{7.98 $\pm$ 5.57} & \multirow{2}{*}{7.23 $\pm$ 3.80} & \multirow{2}{*}{6.64 $\pm$ 3.36} & \multirow{2}{*}{7.04 $\pm$ 3.22} & \multirow{2}{*}{6.68 $\pm$ 2.45} & \multirow{2}{*}{6.71 $\pm$ 3.10} & \multirow{2}{*}{7.47 $\pm$ 2.91} & \multirow{2}{*}{9.04 $\pm$ 4.34} & \multirow{2}{*}{7.35 $\pm$ 3.59} \\
 &  &  &  &  &  &  &  &  &  \\ \hline

\multirow{2}{*}{\shortstack{YOLO-\\Direct\\Regression}} & \multirow{2}{*}{11.20 $\pm$ 9.13} & \multirow{2}{*}{7.89 $\pm$ 4.51} & \multirow{2}{*}{7.10 $\pm$ 3.52} & \multirow{2}{*}{7.69 $\pm$ 3.51} & \multirow{2}{*}{6.97 $\pm$ 2.55} & \multirow{2}{*}{7.90 $\pm$ 4.04} & \multirow{2}{*}{8.26 $\pm$ 3.64} & \multirow{2}{*}{10.71 $\pm$ 6.63} & \multirow{2}{*}{8.47 $\pm$ 4.69} \\
 &  &  &  &  &  &  &  &  &  \\ \hline
 
\multirow{2}{*}{\shortstack{GT-\\Proposed\\Method}} & \multirow{2}{*}{4.51 $\pm$ 3.14} & \multirow{2}{*}{3.89 $\pm$ 1.91} & \multirow{2}{*}{3.62 $\pm$ 1.80} & \multirow{2}{*}{3.79 $\pm$ 1.89} & \multirow{2}{*}{3.63 $\pm$ 1.46} & \multirow{2}{*}{3.4 $\pm$ 1.48} & \multirow{2}{*}{3.64 $\pm$ 1.51} & \multirow{2}{*}{5.68 $\pm$ 3.51} & \multirow{2}{*}{\textbf{4.02 $\mathbf{\pm}$ 2.09}} \\
 &  &  &  &  &  &  &  &  &  \\ \hline

\multirow{2}{*}{\shortstack{YOLO-\\Proposed\\Method}} & \multirow{2}{*}{6.78 $\pm$ 7.37} & \multirow{2}{*}{4.23 $\pm$ 3.00} & \multirow{2}{*}{3.87 $\pm$ 2.05} & \multirow{2}{*}{4.31 $\pm$ 2.43} & \multirow{2}{*}{3.81 $\pm$ 1.64} & \multirow{2}{*}{4.29 $\pm$ 3.54} & \multirow{2}{*}{4.04 $\pm$ 2.03} & \multirow{2}{*}{7.37 $\pm$ 6.67} & \multirow{2}{*}{4.84 $\pm$ 3.59} \\
 &  &  &  &  &  &  &  &  &  \\ \hline
\end{tabular}%
}
\end{table}

\begin{table}[!htbp]
\centering
\caption{Timing analysis of Proposed Method and comparison with other comparing methods}
\vspace{3mm}
\label{tab:time}
\resizebox{416pt}{!}{%
\renewcommand{\arraystretch}{1.8}
\begin{tabular}{|c|c|c|c|c|c|}
\hline
\multirow{2}{*}{Method} & \multirow{2}{*}{\shortstack{Total\\Parameters}} & \multirow{2}{*}{\shortstack{YOLO (ms)}} & \multirow{2}{*}{\shortstack{Fingertip\\Detection (ms)}} & \multirow{2}{*}{\shortstack{Post-\\processing ($\mu$s)}} & \multirow{2}{*}{\shortstack{Total (ms)}} \\
 &  &  &  &  & \\ \hline 

YOLSE & 2,781,669 & 24.00 & 21.82 & 115.25 & 45.94 \\ \hline

Direct Regression & 2,589,775 & 24.00 & 19.78 & 63.95 & 43.84 \\ \hline

Proposed Method & 24,163,654 & 24.00 & 21.99 & 88.44 & 46.08 \\ \hline
\end{tabular}%
}
\end{table}

\subsection{Results}
Table~\ref{tab:classification} shows the results of egocentric gesture recognition in terms of the accuracy, precision, recall, and F1 score of the comparing methods. The overall performance in terms of the mean value of these metrics is also shown in this table. The name of the methods is prefixed by GT as no hand detector is included as preprocessing rather ground truth bounding box is used to directly crop the relevant hand portion from an input image. The results of each method are also presented by including the YOLO hand detector in the first stage, and in this case, the name of the methods is prefixed by YOLO. It can be observed from Table~\ref{tab:classification} that the proposed method has outperformed the other gesture recognition methods and attained very high accuracy in all classes. In particular, the proposed method provides gesture recognition accuracy of at least $99.90\%$ and an F1 score as high as $0.99$.
\par
In estimating the position of fingertips, the distance error between the ground truth coordinate, and the regressed coordinate among the different classes is calculated. Table~\ref{tab:regression} shows the results of the mean and standard deviation of the regression error in pixel (px) for different methods. It is seen from this table that, the proposed fingertip regression approach achieves a better result in terms of the mean and standard deviation of the pixel error as compared to the Direct Regression method, but a comparable performance with the YOLSE method. However, the superiority of the proposed method over the YOLSE method is clear when comparing it with the GT hand image. Nevertheless, the proposed method with the YOLO hand detector has achieved a mean pixel error of $4.84$ px with a standard deviation of $3.59$ px.
\par
In gesture classification, the Direct Regression approach shows a competitive performance with the proposed one. But the mean accuracy, precision, and recall of gesture classification of the proposed method are $1.76\%$, $3.79\%$, and $12.10\%$ higher than the YOLSE method, respectively. On the other hand, in the case of fingertip detection, the YOLSE method shows a competitive performance with the proposed one. However, the mean and standard deviation of the detection error of the Proposed Method is $42.86\%$ and $23.45\%$ less compared to the Direct Regression approach, respectively. Therefore, the Proposed Method is robust both in gesture classification and fingertip detection without any compromise.
\par
\begin{figure}[!t]
\centering 
\subfigure[YOLSE Approach]
{\includegraphics[scale=0.41]{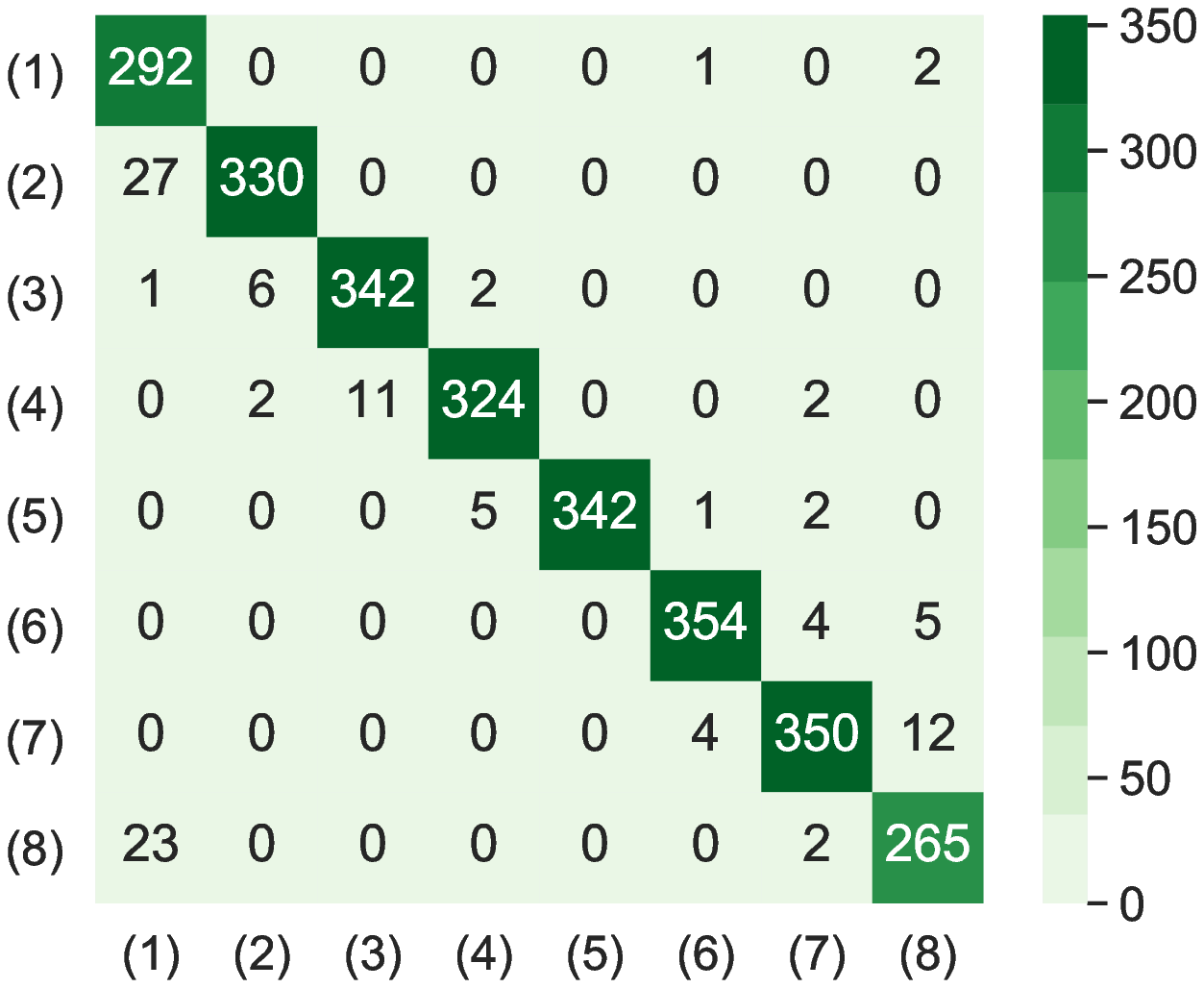}} \hspace{1mm}
\subfigure[Direct Regression Approach]
{\includegraphics[scale=0.41]{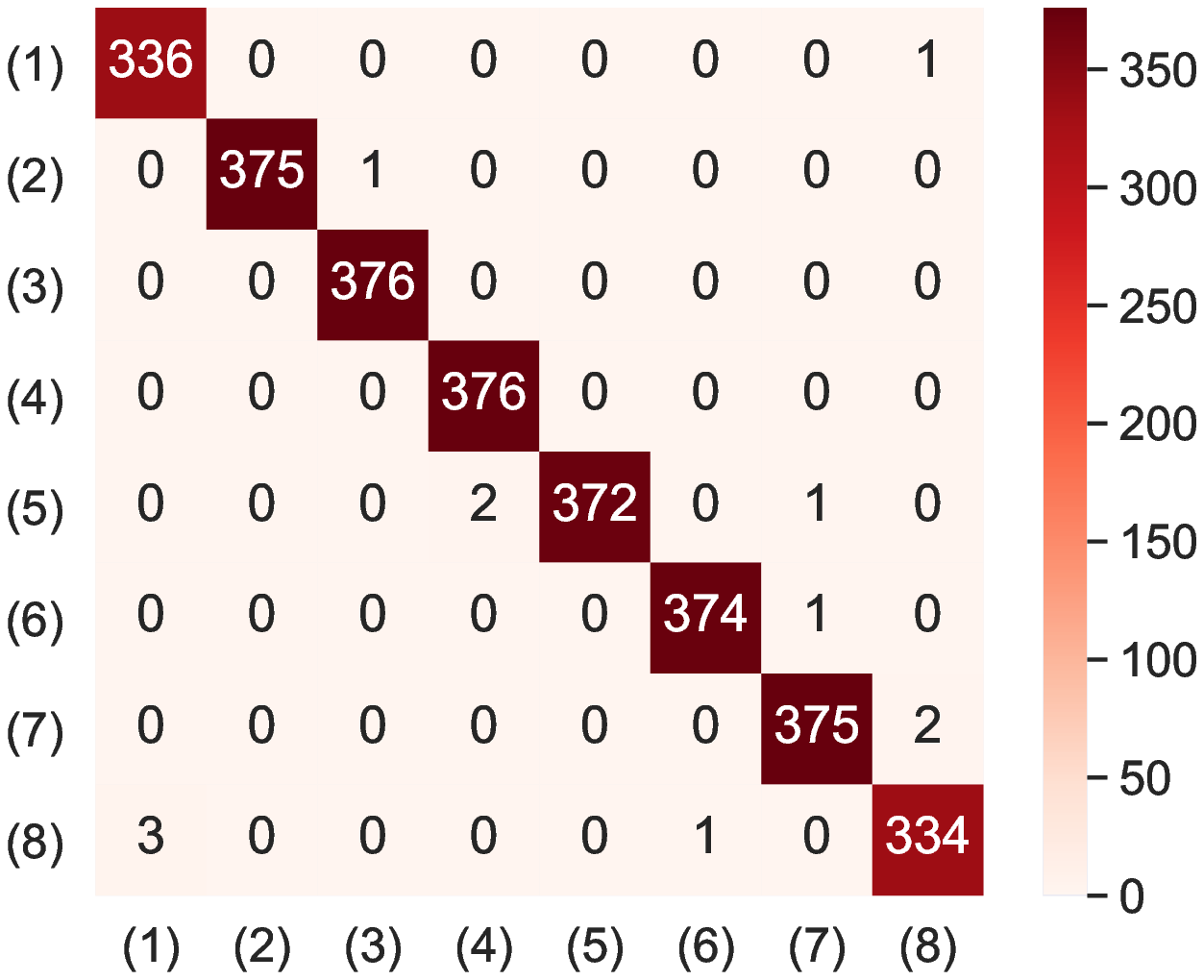}} \hspace{1mm}
\subfigure[Unified Detection Approach]
{\includegraphics[scale=0.41]{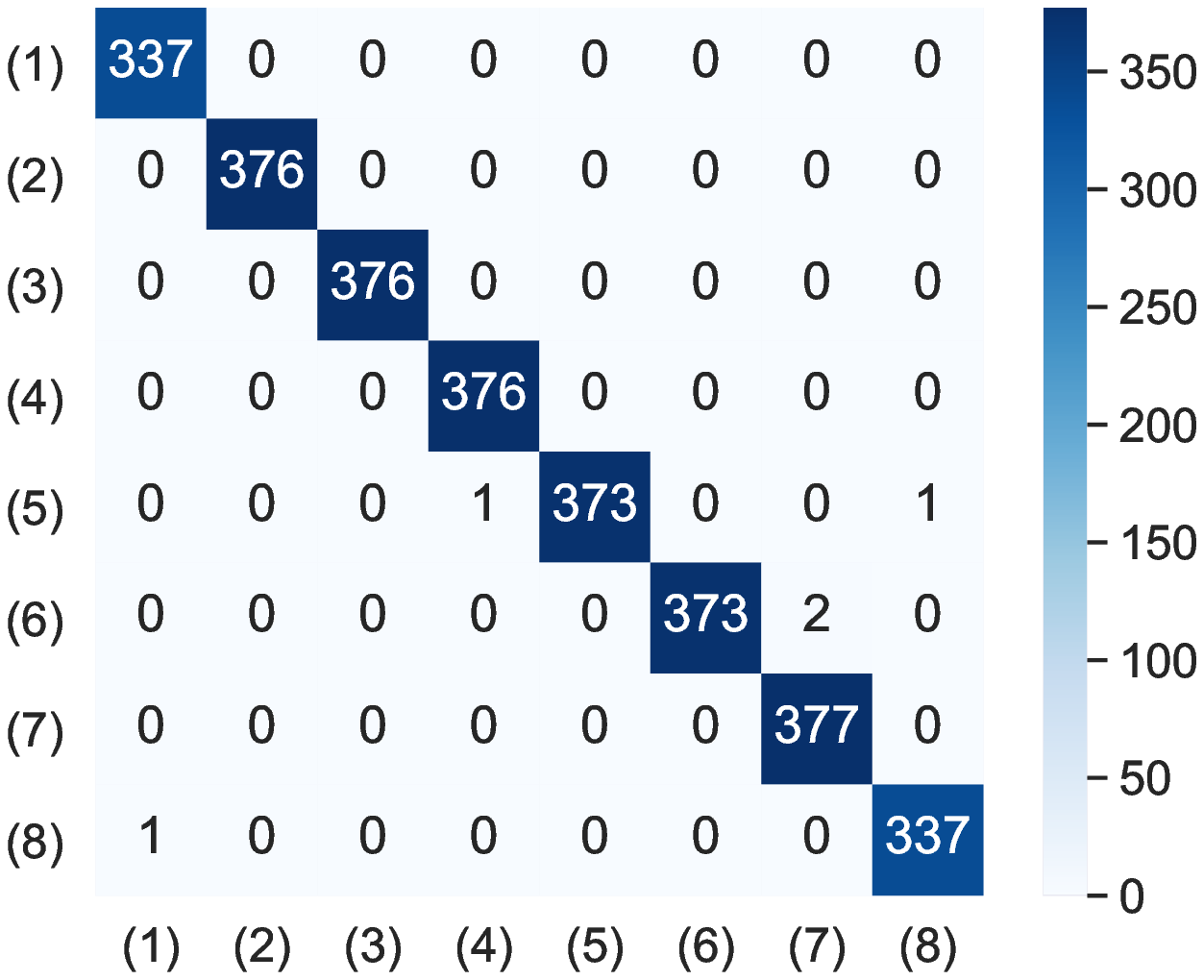}}  \caption{Confusion matrices depicting the performance of the gesture classification by the experimental methods shown in (a) to (c). Here (1) to (8) are representing SingleOne to SingleEight gestures.} 
\label{fig:conf_mat}
\end{figure}

\begin{figure}[!t]
\centerline{\includegraphics[scale=0.582]{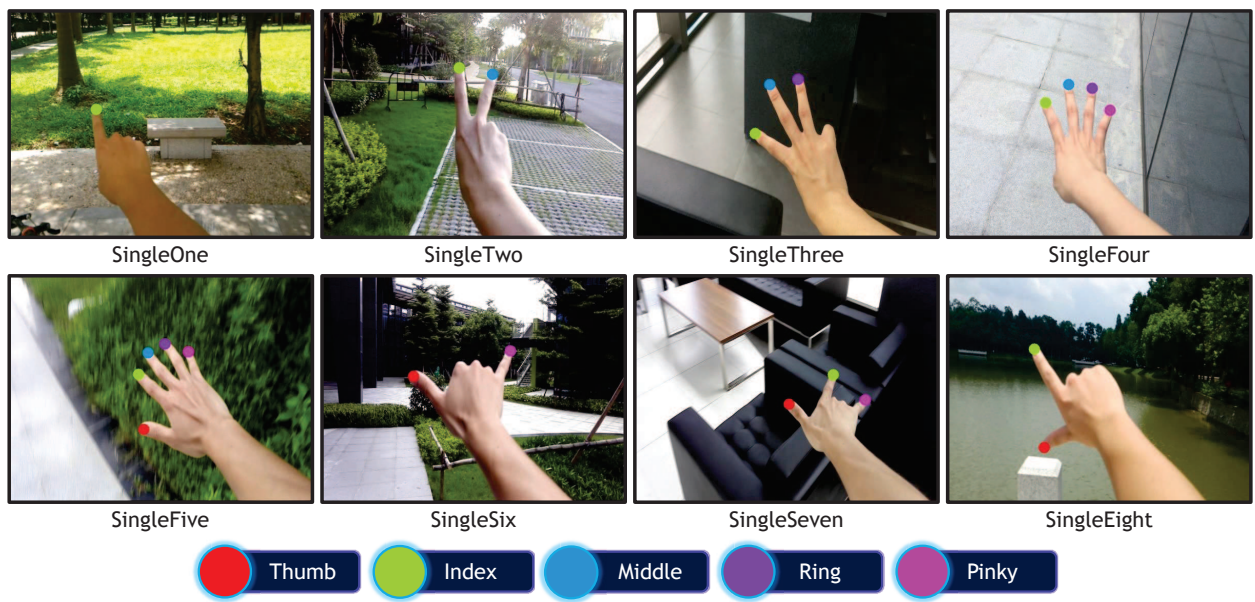}} 
\caption{A visual representation of the outputs of the proposed gesture recognition and fingertip detection model where not only each fingertip is detected but also each finger is classified.}
\label{fig:output}
\end{figure}

 
Figure~\ref{fig:conf_mat} shows the confusion matrices depicting the performance of the classification of gesture by the YOLSE approach, the Direct Regression approach, and the Proposed Method where each row represents the actual class of gesture and each column represents the predicted class of gesture. The figure illustrates that the proposed model has very little confusion in classifying gestures. Figure~\ref{fig:output} shows examples of visual output of the proposed gesture recognition and fingertip detection algorithm of each gesture class where not only each fingertip position is detected but also the type of hand gesture is recognized by classifying each finger. 
\par 
The experiments are performed on a computer with Intel Core i7 CPU with 16 GB memory and NVIDIA RTX2070 Super GPU with 8 GB memory and some of the training portions are conducted using an NVIDIA Titan Xp GPU. The average forward propagation time of the proposed network is $21.99$ ms or $45$ frames per second. Thus, the proposed method satisfies the requirements of real-time implementation. Moreover, a timing analysis of the proposed method and the comparing ones is presented in Table \ref{tab:time}. Although the proposed method has way more parameters compared to the others, the total amount of time required for the proposed method remains almost the same.

\subsection{Ablation Study}
To unfold the full utility and comprehend the contribution of the proposed egocentric gesture recognition and fingertip detection algorithm, we have experimented by alternating and removing different components of the system. The description of the experiments are as follows:

\begin{itemize}
\addtolength{\itemindent}{1mm}

\item[\textbf{(1)}] The proposed method predicts an ensemble of fingertip coordinates and subsequently, ensemble average is taken to predict the final fingertip coordinate. In our first ablation study, we have removed the ensemble averaging from the post-processing stage shown in Figure~\ref{network} and incorporated an averaging layer at the end of the positional output of the network. After the prediction of the ensemble of fingertip coordinates, a Global Average Pooling layer is annexed to take the average which is used as the final output of the fingertip coordinates.
\vskip 1.5mm 

\item[\textbf{(2)}] In our second ablation study, instead of taking an ensemble average of the ensemble of fingertip coordinates, a random sample from the ensemble output is taken. Therefore, we have randomly sampled one of the ten outputs from the ensemble output. That randomly chosen fingertip coordinates are used as the final fingertip coordinates prediction. As we are randomly sampling one output from the ensemble, the post-processing stage is not necessary here.
\vskip 1.5mm 

\item[\textbf{(3)}] In our third ablation study, we have directly regressed the fingertip coordinated from the proposed network. In this experiment, we have removed the Upsampling and Convolution stages after the feature learning stage. The feature learning stage output is used to directly regress the fingertip coordinates using an FC layer with a sigmoid activation function. As we are directly regressing here, the post-processing stage is not necessary here also.
\vskip 1.5mm 
\end{itemize}

\par 

The detection error in terms of pixels (px) of the fingertip coordinates of each class of gestures for the aforementioned studies is presented in Table~\ref{tab:ablation}. In all the cases the ground truth hand bounding box is utilized. The table shows the mean and standard deviation of detection errors of each class as well as for the overall case. In the first study, we have used an averaging layer to take the ensemble average. The averaging layer does not require any parameter to learn, and it is just like averaging in the post-processing stage. Therefore, it is expected to have a similar performance as the proposed method which is apparent from the table. In the second study, we randomly sample fingertip coordinates from the ensemble. Since the ensemble average mitigates the deviation of the prediction from the ground truth value, the output as random samples deviates the performance from the proposed method. In the third study, we directly regressed the fingertip coordinates using an FC layer. The FC layer uses all the features from the previous layer whereas the proposed ensemble output from FCN uses different input features of the previous layer. Therefore, the difference of errors between the results of the direct regression approach and that of the proposed method is expected as shown in Table~\ref{tab:ablation}.

\begin{table}[!t]
\centering
\caption{Results of the ablation study of the proposed egocentric hand gesture recognition and fingertip detection algorithm}
\vspace{3mm}
\label{tab:ablation}
\resizebox{\textwidth}{!}{%
\renewcommand{\arraystretch}{1.8}
\begin{tabular}{|c|c|c|c|c|c|c|c|c|c|}
\hline
\multirow{2}{*}{Method} & \multicolumn{8}{c|}{Gesture} & \multirow{2}{*}{\shortstack{Overall\\Error\\(px)}} \\ \cline{2-9}
 & SingleOne & SingleTwo & SingleThree & SingleFour & SingleFive & SingleSix & SingleSeven & SingleEight &  \\ \hline

\multirow{3}{*}{\shortstack{(1)~Method with \\ an Averaging Layer \\ Included in the CNN \\Architecture}} & \multirow{3}{*}{4.64 $\pm$ 3.17} & \multirow{3}{*}{3.94 $\pm$ 1.86} & \multirow{3}{*}{3.61 $\pm$ 1.86} & \multirow{3}{*}{3.85 $\pm$ 1.95} & \multirow{3}{*}{3.63 $\pm$ 1.53} & \multirow{3}{*}{3.51 $\pm$ 1.5} & \multirow{3}{*}{3.65 $\pm$ 1.46} & \multirow{3}{*}{5.8 $\pm$ 3.77} & \multirow{3}{*}{4.08 $\pm$ 2.14} \\
 &  &  &  &  &  &  &  &  &  \\
 &  &  &  &  &  &  &  &  &  \\ \hline

\multirow{3}{*}{\shortstack{(2)~Method with \\Randomly Sampled \\Output from the \\Ensemble}} & \multirow{3}{*}{5.21 $\pm$ 3.43} & \multirow{3}{*}{4.57 $\pm$ 2.3} & \multirow{3}{*}{4.24 $\pm$ 2.23} & \multirow{3}{*}{4.28 $\pm$ 1.98} & \multirow{3}{*}{4.17 $\pm$ 1.87} & \multirow{3}{*}{3.98 $\pm$ 1.78} & \multirow{3}{*}{4.23 $\pm$ 1.68} & \multirow{3}{*}{6.41 $\pm$ 3.56} & \multirow{3}{*}{4.63 $\pm$ 2.35} \\
 &  &  &  &  &  &  &  &  &  \\
 &  &  &  &  &  &  &  &  &  \\ \hline

\multirow{3}{*}{\shortstack{(3)~Method with \\Direct Regression \\of Fingertip \\Coordinates}} & \multirow{3}{*}{5.97 $\pm$ 3.75} & \multirow{3}{*}{6.23 $\pm$ 2.21} & \multirow{3}{*}{6.01 $\pm$ 2.49} & \multirow{3}{*}{5.94 $\pm$ 2.11} & \multirow{3}{*}{5.12 $\pm$ 1.66} & \multirow{3}{*}{5.45 $\pm$ 2.32} & \multirow{3}{*}{4.88 $\pm$ 2.25} & \multirow{3}{*}{6.88 $\pm$ 3.29} & \multirow{3}{*}{5.81 $\pm$ 2.51} \\
 &  &  &  &  &  &  &  &  &  \\
 &  &  &  &  &  &  &  &  &  \\ \hline
 
\multirow{2}{*}{\shortstack{Proposed Method}} & \multirow{2}{*}{4.51 $\pm$ 3.14} & \multirow{2}{*}{3.89 $\pm$ 1.91} & \multirow{2}{*}{3.62 $\pm$ 1.80} & \multirow{2}{*}{3.79 $\pm$ 1.89} & \multirow{2}{*}{3.63 $\pm$ 1.46} & \multirow{2}{*}{3.4 $\pm$ 1.48} & \multirow{2}{*}{3.64 $\pm$ 1.51} & \multirow{2}{*}{5.68 $\pm$ 3.51} & \multirow{2}{*}{\textbf{4.02 $\mathbf{\pm}$ 2.09}} \\
 &  &  &  &  &  &  &  &  &  \\ \hline

\end{tabular}%
}
\end{table}

\section{Detection In-The-Wild and Application In VR}\label{wild}
\begin{figure}[!t]
\centerline{\includegraphics[scale=0.556]{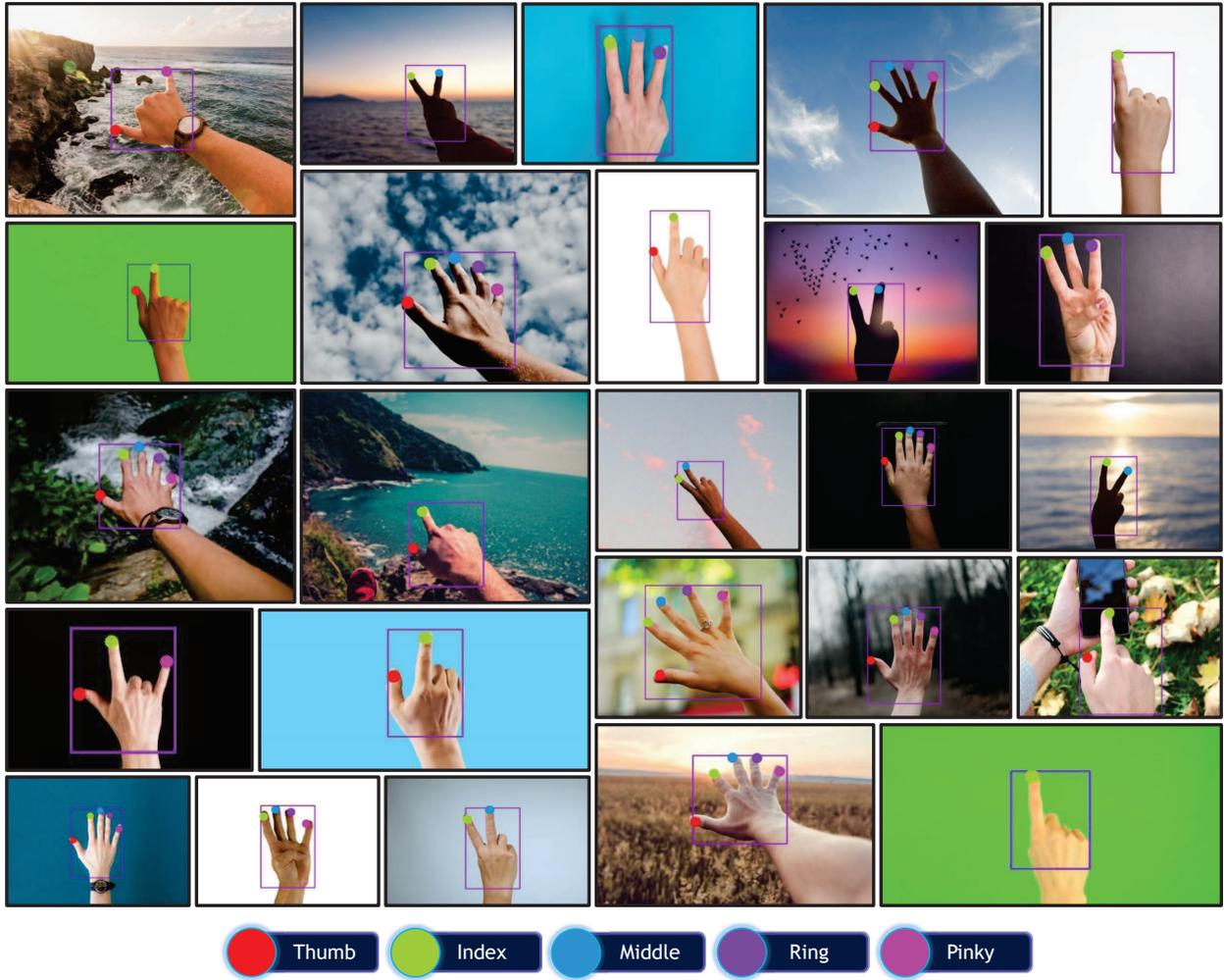}}
\caption{Prediction of the model using random images collected over the internet to show the real-life usability of the proposed method.}
\label{fig:wild}
\end{figure}

To evaluate the performance of the proposed method in real-life scenarios, $25$ publicly available hand gesture images are collected from the internet. The imaging conditions of this wild set of gesture images are quite different as compared to the SCUT-Ego-Gesture database. In particular, they are different in terms of background, illumination, resolution, and pose of the fingers. Moreover, the hand shape and the color in these images differ from the SCUT-Ego-Gesture database. Fig. \ref{fig:wild} shows the output images with the prediction of the proposed method. It is seen from the output images that the proposed method is capable of successfully predicting all the gestures and detects all the fingertips of the images collected from the internet despite being different from the database to a large extent.

\begin{figure}[!t]
\centerline{\includegraphics[scale=0.54]{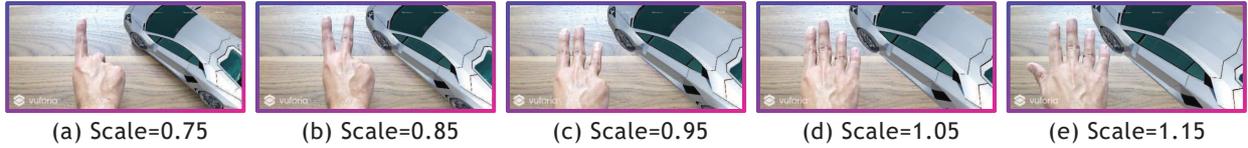}}
\caption{VR application of the proposed egocentric gesture recognition and fingertip detection algorithm. In this experimental demo application, the scale of the virtual 3D object, i.e. car, is modified depending the number of fingers in a gesture shown in from (a) to (e) where the scale is incremented from $0.75$ to $1.15$.}
\label{fig:vr}
\end{figure}

To show the real-life feasibility of the proposed method in VR applications, we have also demonstrated a proof-of-concept VR application. In this demonstration, we have placed a virtual 3D object, e.g., car, on the surface and modified its scale based on the number of fingers in the egocentric gesture. The initial scale of the virtual 3D object is set as $0.75$ which is for one finger only and the scale is set as $1.15$ for five fingers. The scale value is incremented by $0.10$ for each finger. Figure~\ref{fig:vr} shows an illustration of the VR application where the scale of the virtual object is incremented with the number of fingers. Therefore, in real-life HCI, VR, and MR applications, the proposed method can play an indispensable role.

\section{Conclusion}\label{sec:conclusions}
In this paper, a new CNN-based method is proposed that unifies the gesture recognition and prediction of fingertip position for \text{egocentric vision} in a single step process. In particular, the proposed method regressed the ensemble of the position of fingertips using a fully convolutional network instead of directly regressing the positions of fingertips using the fully connected layer. The experiments have been carried out by employing a commonly referred SCUT-Ego-Gesture database. The accuracy of the automatic gesture recognition has been found to be at least $99.90\%$, and the minimum F1 score among the classes has been found to be at least $0.9955$. The mean pixel error in fingertip detection among the classes has been found to be $4.84$ px. As the proposed method uses a single network for both gesture recognition and fingertip detection, it is very fast and meets the requirements of real-time implementation. The proposed method has achieved lower false positive and false negative rates in classification and made less localization error in regression as compared to the previous direct regression and Heatmap-based YOLSE methods. Moreover, we have presented an ablation study by altering different parts of the proposed algorithm to demonstrate the effectiveness of the methodology. The performance of the proposed method is also ensured by experimentation using the hand gesture images available in-the-wild and by using an application setting in the VR environment. In the VR application, we have successfully modified the scale of a virtual 3D car. The method in its current form uses a relatively large network for feature extraction. Even though, the system can perform in real-time, constructing a smaller architecture without sacrificing accuracy will benefit the method to become even faster. Secondly, the datasets for egocentric vision must be improved significantly. While various datasets are available online, one major bottleneck of these datasets is that they either do not consider different skin color and age of the users or they do not consider various background and lighting condition. A dataset that takes these aspects into account will be very beneficial to the field. Thirdly, we only focus on one hand in the egocentric vision in this paper. Future works would investigate incorporating both hands into the pipeline. In view of the overall fingertip detection performance of the proposed method as verified through experimentations in the egocentric hand vision and its prospective improvement strategies, we expect that the proposed method can play a significant role in the HCI, VR, and MR applications.

\section{CRediT Authorship Contribution Statement}
\textbf{Mohammad Mahmudul Alam:} Conceptualization, Methodology, Software, Formal analysis, Data Curation, Writing - Original draft, Writing - Review \& Editing. \textbf{Mohammad Tariqul Islam:} Validation, Formal analysis, Writing - Original Draft. \textbf{S. M. Mahbubur Rahman:} Formal analysis, Resources, Writing - Original Draft, Writing - Review \& Editing, Supervision.

\section*{Acknowledgment}
The authors gratefully acknowledge the support of NVIDIA Corporation for the donation of a Titan Xp GPU that was used in this research. The authors also would like to thank the anonymous reviewers for their valuable comments, which have been useful in improving the quality of the paper.

\bibliographystyle{elsarticle-num}

\end{document}